\documentclass[journal]{IEEEtran}

\usepackage{url}
\usepackage{bbm}
\usepackage{soul}
\usepackage{xcolor}
\usepackage{framed}
\usepackage{colortbl}
\usepackage{array}
\usepackage{mdwtab}
\usepackage{mdwmath}
\usepackage{eqparbox}
\usepackage{amsfonts}
\usepackage{booktabs}
\colorlet{shadecolor}{yellow}
\usepackage[pdftex]{graphicx}
\usepackage[table]{xcolor}
\usepackage[cmex10]{amsmath}
\usepackage[citecolor=blue,colorlinks]{hyperref}

\definecolor{lightyellow}{RGB}{255, 255, 204}
\definecolor{lightorange}{RGB}{255, 204, 153}

\newcommand{\revise}[1]{{\color{black}{#1}}}

\begin{document}

\title{FlexPara: Flexible Neural Surface Parameterization}
\author{Yuming Zhao, Qijian Zhang, Junhui Hou, Jiazhi Xia, Wenping Wang, and Ying He 

\thanks{This work was supported in part by the NSFC Excellent Young Scientists Fund 62422118 and U23A20313, in part by the Hong Kong Research Grants Council under Grants 11219324 and 11219422, and in part by the Science Foundation for Distinguished Young Scholars of Hunan Province (NO. 2023JJ10080).(\textit{Corresponding author: Junhui Hou})}
\thanks{Y. Zhao and J. Hou are with the Department of Computer Science, City University of Hong Kong, Hong Kong SAR. Email: yumzhao2-c@my.cityu.edu.hk; jh.hou@cityu.edu.hk.}
\thanks{Q. Zhang is with the TiMi L1 Studio of Tencent Games, China. Email: keeganzhang@tencent.com.}
\thanks{J. Xia is with the School of Computer Science and Engineering, Central South University, China. Email: xiajiazhi@csu.edu.cn}
\thanks{W. Wang is with the Department of Computer Science \& Engineering, Texas A\&M University, USA. Email: wenping@tamu.edu.}
\thanks{Y. He is with the College of Computing and Data Science, Nanyang Technological University, Singapore. Email:yhe@ntu.edu.sg}
}

\markboth{Revised Manuscript submitted to IEEE TPAMI}{}

\maketitle

\begin{abstract}
Surface parameterization is a fundamental geometry processing task, laying the foundations for the visual presentation of 3D assets and numerous downstream shape analysis scenarios. Conventional parameterization approaches demand high-quality mesh triangulation and are restricted to certain simple topologies unless additional surface cutting and decomposition are provided. In practice, the optimal configurations (e.g., type of parameterization domains, distribution of cutting seams, number of mapping charts) may vary drastically with different surface structures and task characteristics, thus requiring more flexible and controllable processing pipelines. To this end, this paper introduces FlexPara, an unsupervised neural optimization framework to achieve both global and multi-chart surface parameterizations by establishing point-wise mappings between 3D surface points and adaptively-deformed 2D UV coordinates. We ingeniously design and combine a series of geometrically-interpretable sub-networks, with specific functionalities of cutting, deforming, unwrapping, and wrapping, to construct a bi-directional cycle mapping framework for global parameterization without the need for manually specified cutting seams. Furthermore, we construct a multi-chart parameterization framework with adaptively-learned chart assignment. Extensive experiments demonstrate the universality, superiority, and inspiring potential of our neural surface parameterization paradigm.  The code will be publicly available at \href{https://github.com/AidenZhao/FlexPara}{https://github.com/AidenZhao/FlexPara}.
\end{abstract}


\IEEEpeerreviewmaketitle

\section{Introduction}
\begin{figure*}[!t]
    \centering
    \includegraphics[width=1.0\linewidth]{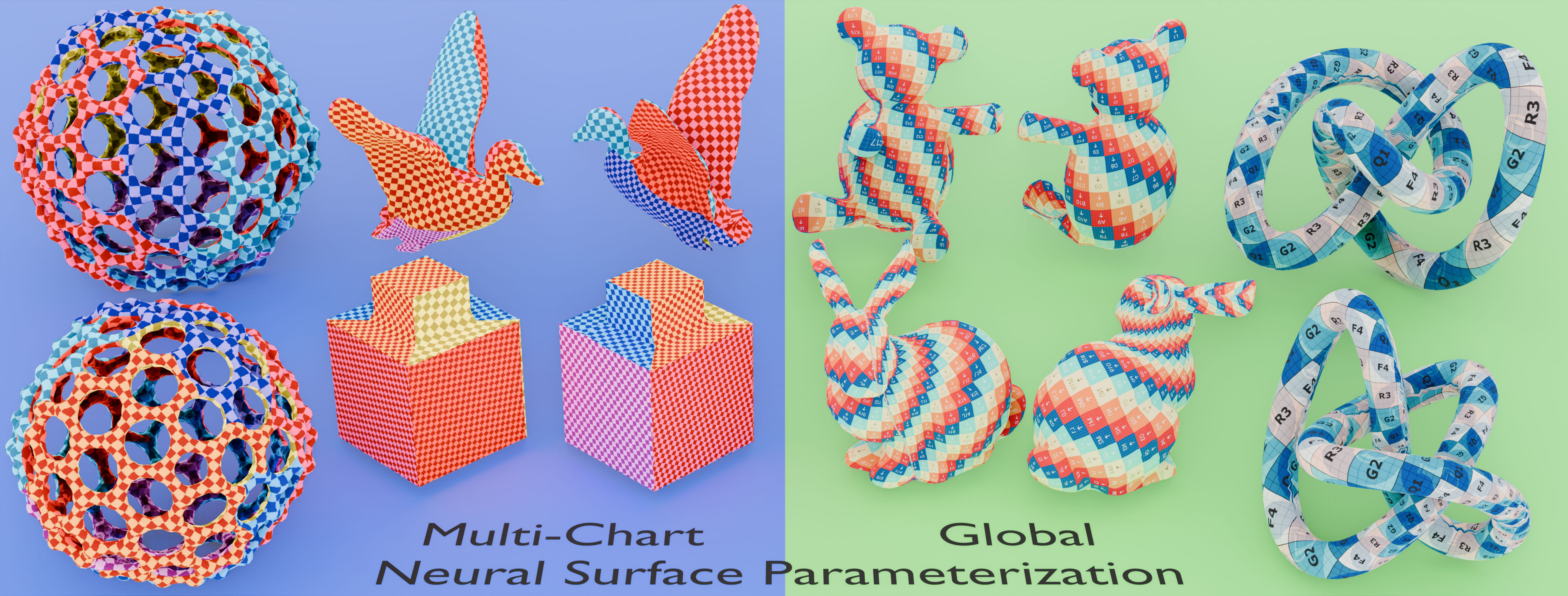} \vspace{-0.2cm}
    \caption{Examples of global and multi-chart parameterizations achieved by our methods using grid-checkerboard texture mapping to visualize parametric distortion. We illustrate global parameterization on the Stanford Bunny, Bear and a twisted knot model. For multi-chart parameterization, we present results on a polycube model with sharp features, a bird model exhibiting symmetry and thin structures, and a spherical gyroid model with high genus. Note that our multi-chart parameterization effectively preserves sharp features and global symmetry, and robustly handles models with intricate topologies. The grid-checkerboard texture highlights low angular and area distortions achieved by both our global and multi-chart parameterization algorithms. }
    \label{teaser image}
\end{figure*}

Surface parameterization, commonly referred to as UV unwrapping, is a cornerstone of computer graphics and geometry processing. This technique involves flattening a 3D geometric surface onto a 2D plane, known as the parameter domain. Specifically, for any 3D point $(x, y, z)$ on the surface, a corresponding 2D coordinate $(u,v)$ is defined, subject to constraints on continuity and distortion. Such point-to-point mappings are essential in modern graphics rendering pipelines, particularly for texture mapping, and are widely applied in downstream geometry processing tasks, including remeshing, mesh completion, mesh compression, detail transfer, surface fitting, and editing. The quality of surface parameterization significantly influences the performance of these tasks, making the development of efficient and accurate parameterization techniques a vital research focus in computer graphics and related disciplines.

Theoretically, given any two geometric surfaces with identical/similar topological structures, there exists a bijective mapping between them. Nevertheless, when the topology of the target 3D surface becomes complicated (e.g., with high genus), one must pre-open the original mesh to a sufficiently-developable disk along appropriate cutting seams. Consequently, the current industrial practice for implementing UV unwrapping is typically composed of two stages: (1) manually specifying some necessary cutting seams on the original mesh; (2) applying mature disk-topology-oriented parameterization algorithms (such as LSCM \cite{levy2002least} and ABF \cite{sheffer2001parameterization,sheffer2005abf++}) which have been well integrated into popular 3D modeling software (e.g., Blender, Unity, 3Ds Max, etc.) to produce per-vertex 2D UV coordinates. In practice, such a stage-wise UV unwrapping pipeline still shows the following limitations and inconveniences: 

\begin{enumerate}

\item Commonly-used surface parameterization algorithms are designed for well-behaved meshes, which are typically produced by specialized 3D modelers and technical artists. However, with the advent of user-generated content fueled by the rapidly growing 3D sensing, reconstruction \cite{mildenhall2021nerf,kerbl20233d}, and generation \cite{nichol2022point,poole2023dreamfusion,lin2023magic3d,siddiqui2024meshgpt} techniques, there emerges an urgent need for dealing with ordinary 3D data possibly with unruly anomalies, inferior triangulations, and non-ideal geometries.
    
\item Despite the existence of a few early attempts at heuristic seam generation \cite{sheffer2002seamster,erickson2002optimally}, the process of finding high-quality cutting seams in practical applications still relies on manual efforts and personal experience. Hence, the entire workflow remains subjective and semi-automated, leading to reduced reliability and efficiency, especially when dealing with complex 3D models that users are unfamiliar with. Besides, since cutting is actually achieved via edge selection, one may need to repeatedly adjust the distribution of mesh edges to allow the cutting seams to walk through.
    
\item The procedures of cutting the original mesh into a disk and then flattening the resulting disk onto the parameter domain should have been mutually influenced and jointly optimized; otherwise, the overall surface parameterization results could be sub-optimal.  

\end{enumerate}

In recent years, there has emerged a new family of neural parameterization approaches targeted at learning parameterized 3D geometric representations via neural network architectures. The pioneering works of FoldingNet \cite{yang2018foldingnet} and AtlasNet \cite{groueix2018papier} are among the best two representatives, which can build point-wise mappings via deforming a pre-defined 2D lattice grid to reconstruct the target 3D shape. RegGeoNet \cite{zhang2022reggeonet} and Flattening-Net \cite{zhang2023flattening} tend to achieve fixed-boundary rectangular structurization of irregular 3D point clouds through geometry image \cite{gu2002geometry,losasso2003smooth} representations. However, these two approaches cannot be regarded as real-sense surface parameterization due to their lack of mapping constraints. DiffSR \cite{bednarik2020shape} and Nuvo \cite{srinivasan2025nuvo} focus on the local parameterization of the original 3D surface but with explicit and stronger constraints on the learned neural mapping process. Their difference lies in that DiffSR aggregates multi-patch parameterizations to reconstruct the original geometric surface, while Nuvo adaptively assigns surface points to different charts in a probabilistic manner.  

In this paper, we make the first attempt to investigate neural surface parameterization featured by both global mapping and free boundary. We introduce a universal and fully-automated UV unwrapping approach to build point-to-point mappings between 3D points lying on the target geometric surface and 2D UV coordinates within the adaptively-deformed parameter domain. In general, our approach is designed as an unsupervised learning pipeline (i.e., no ground-truth UV maps are collected), running in a per-model overfitting manner. To mimic the actual physical process, we ingeniously design a series of geometrically-meaningful sub-networks with specific functionalities of surface cutting, UV deforming, 3D-to-2D unwrapping, and 2D-to-3D wrapping, which are further assembled into a bi-directional cycle mapping framework. By optimizing a combination of different loss functions and auxiliary differential geometric constraints, our approach is able to find reasonable 3D cutting seams and 2D UV boundaries, leading to superior parameterization quality when dealing with different degrees of geometric and topological complexities. 

Additionally, based on the global parameterization framework, we have designed a joint multi-chart parameterization framework specifically. At the front end of this framework, we have developed a differentiable chart assignment score module that predicts the segmentation of the entire model into charts. Concurrently, we introduce the single-directional cycle mapping structure in parallel at the back end to perform local flattening for each chart. Compared to the global parameterization task, the local parameterization for segmented charts is relatively simpler. Therefore, we opted to use single-directional cycle mapping to enhance operational efficiency.

Comprehensive experiments demonstrate the advantages of our approach over traditional state-of-the-art approaches. Figure \ref{teaser image} shows some typical examples achieved by our methods with grid-checkboard texture. In summary, the main contributions of our research are as follows.

\begin{itemize}

\item We propose a global neural parameterization method that employs a bi-directional cycle mapping to simultaneously learn surface cutting and parameterization mapping, ensuring precise seam placement and UV mapping. This approach is applicable to geometric surfaces with diverse topologies.

\item Building on global parameterization, we introduce a multi-chart parameterization approach. This method leverages a differentiable chart assignment module for chart segmentation and allocation, combined with single-directional cycle mappings for UV mapping of individual charts. Through end-to-end joint optimization, it ensures optimal chart segmentation and UV mapping.

\end{itemize}

This manuscript presents substantial extensions and improvements over our earlier conference version \cite{zhangflatten}. The key additions are as follows:

\begin{itemize}

\item In the context of global parameterization, we introduce a novel composite distortion loss constraint, which significantly improves the robustness and quality of global parameterization across diverse input geometries.

\item Building upon our global surface neural parameterization framework, we introduce a new multi-chart extension that produces low-distortion UV mappings while maintaining a controlled number of charts, striking a balance between quality and practical applicability. 

\item We expand the experimental evaluation to a broader dataset comprising challenging models with diverse geometric features and complex topologies. Our method is benchmarked against recent neural parameterization approaches, joint seam-UV optimization methods, and traditional methods that rely on manually specified seams. These comparisons validate the effectiveness of our framework. The results validate the effectiveness of our framework, demonstrating its ability to achieve low distortion while maintaining a limited number of charts and constrained seam lengths. Notably, our method consistently outperforms existing techniques on high-genus and topologically complex models, underscoring its robustness and scalability in challenging scenarios.

\end{itemize}

The remainder of the paper is organized as follows. Section \ref{Related Work} reviews existing works most related to this work. In Section \ref{sec:method}, we introduce the overall architecture of our FlexPara, a flexible unsupervised neural surface parameterization framework, which unifies free-boundary global (single-chart) and local (multi-chart) parameterizations. In Section \ref{Experiment}, we validate the effectiveness of our proposed method on 3D geometry in different types. Finally, Section \ref{Conclusion} concludes this paper.

\section{Related Work} \label{Related Work}

\subsection{Traditional Surface Parameterization Approaches}

Within the computer graphics and geometry processing communities, mesh parameterization \cite{floater2005surface,sheffer2007mesh} is a fundamental and long-standing task, which has been extensively investigated owing to its diverse downstream applications, such as remeshing, morphing, editing, texture mapping, and compression. Early approaches typically formulate this task as a Laplacian problem, where boundary points are anchored to a pre-determined convex 2D curve \cite{eck1995multiresolution,floater1997parametrization}, tackled by sparse linear system solvers. Such a linear strategy is appreciated for its simplicity, efficiency, and the guarantee of bijectivity. However, the rigidity of fixing boundary points in the parameter domain usually leads to significant distortions. In response to these issues, free-boundary parameterization algorithms \cite{sheffer2005abf++,liu2008local} have been proposed, offering a more flexible setting by relaxing boundary constraints. Although these approaches bring certain improvements, they often struggle to maintain global bijectivity. More recent state-of-the-art approaches \cite{rabinovich2017scalable,smith2015bijective} shift towards minimizing simpler proxy energies by alternating between local and global optimization phases, which not only accelerate convergence but also enhance the overall parameterization quality. Despite continuous advancements, the working mechanisms of all these approaches fundamentally rely on the connectivity information inherent in mesh structures, casting doubt on their applicability to unstructured point clouds. Moreover, mesh-oriented parameterization approaches typically assume that the inputs are well-behaved meshes possessing relatively regular triangulations. When faced with input meshes of inferior quality characterized by irregular triangulations and anomalies, remeshing becomes an indispensable procedure. Additionally, how to deal with non-ideal geometries and complex topologies (e.g., non-manifolds, multiply-connected components, thin structures) has always been highly challenging.  Unlike surface meshes, unstructured point clouds lack information regarding the connectivity between points, and thus are much more difficult to parameterize. Hence, only a relatively limited number of prior works \cite{zhu2022review} have focused on the parameterization of surface point clouds. Early studies investigate various specialized parameterization strategies for disk-type \cite{floater2001meshless,azariadis2007product,zhang2010mesh}, genus-0 \cite{zwicker2004meshing,liang2012geometric}, genus-1 \cite{tewari2006meshing}, well-sampled \cite{guo2006meshless} or low-quality \cite{meng2013parameterization} point clouds. Later approaches achieve fixed-boundary spherical/rectangular parameterization through approximating the Laplace-Beltrami operators \cite{choi2016spherical} or Teichmüller extremal mappings \cite{meng2016tempo}. More recently, FBCP-PC \cite{choi2022free} further pursues free-boundary conformal parameterization, in which a new point cloud Laplacian approximation scheme is proposed for handling boundary non-convexity.

\begin{figure*}[!t]
	\centering
	\includegraphics[width=1.0\linewidth]{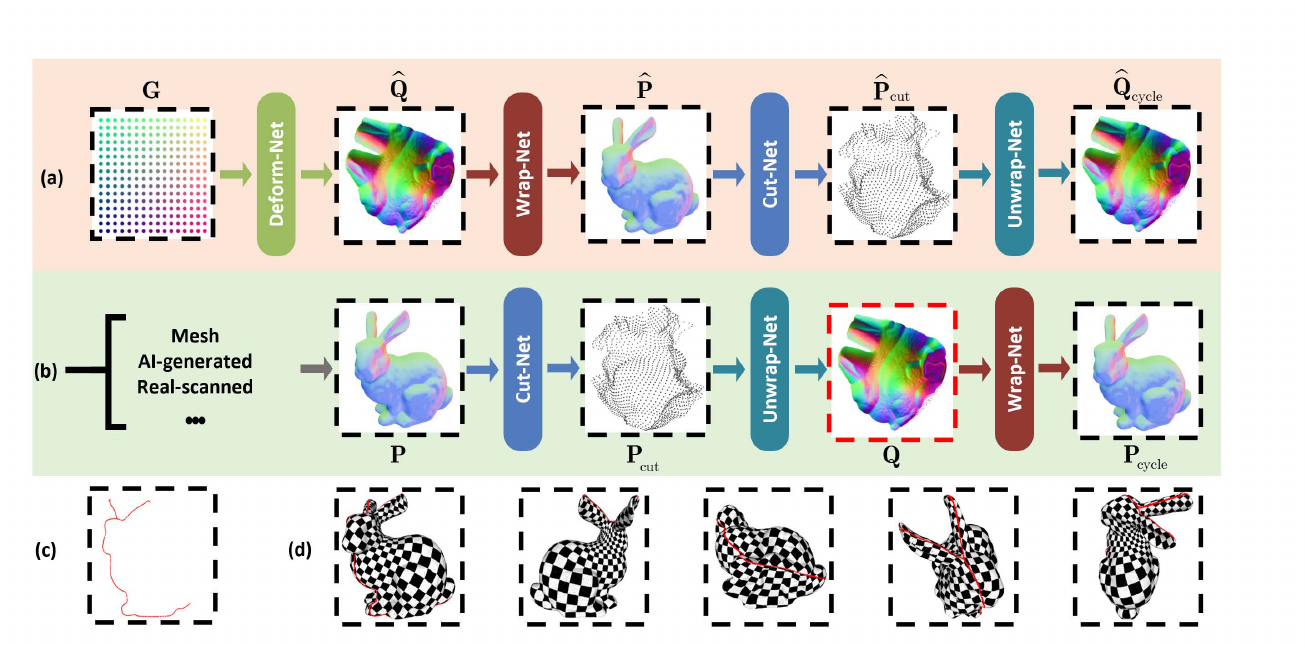} 
	\caption{Illustration of bi-directional cycle mapping for global neural surface parameterization, composed of (a) 2D$\rightarrow$3D$\rightarrow$2D cycle mapping branch, and (b) 3D$\rightarrow$2D$\rightarrow$3D cycle mapping branch. Modules with the same color share network parameters. (c) shows the learned cutting seams. (d) shows the grid-checkboard image texture mapping.}
	\label{pipline_global}
\end{figure*}
\subsection{Neural Surface Parameterization Approaches}

Driven by the remarkable success of deep learning, there is a family of recent works applying neural networks to learn parameterized 3D geometric representations. FoldingNet \cite{yang2018foldingnet} proposes to deform a uniform 2D grid to reconstruct the target 3D point cloud for unsupervised geometric feature learning. AtlasNet \cite{groueix2018papier} applies multi-grid deformation to learn locally parameterized representations. Subsequently, a series of follow-up research inherits such a ``folding-style'' parameterization paradigm as pioneered by \cite{yang2018foldingnet,groueix2018papier} and investigates different aspects of modifications. GTIF \cite{chen2019deep} introduces graph topology inference and filtering mechanisms, empowering the decoder to preserve more representative geometric features in the latent space. EleStruc \cite{deprelle2019learning} proposes to perform shape reconstruction from learnable 3D elementary structures, rather than a pre-defined 2D lattice. Similarly, TearingNet \cite{pang2021tearingnet} adaptively breaks the edges of an initial primitive graph for emulating the topology of the target 3D point cloud, which can effectively deal with higher-genus or multi-object inputs. In fact, the ultimate goal of the above line of approaches is to learn expressive shape codewords by means of deformation-driven 3D surface reconstruction. The characteristics of surface parameterization, i.e., the mapping process between 3D surfaces and 2D parameter domains, are barely considered.

In contrast to the extensive research in the fields of deep learning-based 3D geometric reconstruction and feature learning, there only exist a few studies that particularly focus on neural surface parameterization. DiffSR \cite{bednarik2020shape} adopts the basic multi-patch reconstruction framework \cite{groueix2018papier} and explicitly regularizes multiple differential surface properties. NSM \cite{morreale2021neural} explores neural encoding of surface maps by overfitting a neural network to an existing UV parameterization pre-computed via standard mesh parameterization algorithms \cite{tutte1963draw,rabinovich2017scalable}. DA-Wand \cite{liu2023wand} constructs a parameterization-oriented mesh segmentation framework. Around a specified triangle, it learns to select a local sub-region, which is supposed to be sufficiently developable to produce a low-distortion parameterization. Inheriting a geometry image~\cite{gu2002geometry} representation paradigm, RegGeoNet \cite{zhang2022reggeonet}, Flattening-Net \cite{zhang2023flattening}, and SPCV \cite{zeng2024dynamic} learn deep regular representations of unstructured 3D point clouds. However, these approaches lack explicit constraints on the parameterization distortions, and their local-to-global assembling procedures are hard to control. More recently, Nuvo \cite{srinivasan2025nuvo} proposes a neural UV mapping framework that operates on oriented 3D points sampled from arbitrary 3D representations, liberating from the stringent quality demands of mesh triangulation. This approach assigns the original surface to multiple charts and ignores the packing procedure, thus essentially differing from our targeted global parameterization setting.

\section{Proposed Method} \label{sec:method}

We propose FlexPara, a flexible unsupervised neural 3D surface parameterization framework, which unifies free-boundary global (single-chart) and local (multi-chart) parameterizations without the need for additionally specifying high-quality cutting seams and surface decompositions.

In what follows, we initially introduce a bi-directional cycle mapping framework for achieving \textit{global} surface parameterization in Section~\ref{sec:method-global-para}. By integrating adaptive chart assignment, we further build a unified \textit{multi-chart} surface parameterization framework in Section~\ref{sec:method-local-para}.

\subsection{Global Neural Surface Parameterization} \label{sec:method-global-para}

Given a 3D mesh $\mathcal{X}$ comprising vertex positions $\mathbf{P} \in \mathbb{R}^{V \times 3}$, triangular faces $\mathbf{F} \in \mathbb{R}^{T \times 3}$, and per-vertex normals $\mathbf{N} \in \mathbb{R}^{V \times 3}$ (which can be conveniently computed from $\mathbf{P}$ and $\mathbf{F}$), we aim to point-wisely map $\mathbf{P}$ onto the 2D planar domain, producing UV coordinates $\mathbf{Q} \in \mathbb{R}^{V\times2}$, in which $\mathbf{P}$ and $\mathbf{Q}$ are row-wisely corresponded.

The most straightforward way of building a learning-based surface parameterization framework would be explicitly supervising the generation of the desired $\mathbf{Q}$, which can be hindered by the lack of high-quality UV unwrapping results as ground-truth training data.

Another feasible framework is to subtly choose an opposite mapping direction by wrapping a set of adaptively-deformable 2D points $\mathbf{\hat{Q}} \in \mathbb{R}^{V\times2}$ (which can be considered as potentially-optimal UV coordinates) onto the target 3D geometric surface. Specifically, treating $\mathbf{\hat{Q}}$ as an unstructured point set, we point-wisely transform $\mathbf{\hat{Q}}$ to 3D space to produce a 3D point cloud $\mathbf{\hat{P}} \in \mathbb{R}^{V \times 3}$, such that $\mathbf{\hat{Q}}$ and $\mathbf{\hat{P}}$ are row-wisely corresponded, followed by minimizing certain point set similarity measurements (e.g., Chamfer Distance) between the generated $\mathbf{\hat{P}}$ and the original surface points $\mathbf{P}$. Under ideal situations, once the two 3D point sets $\mathbf{\hat{P}}$ and $\mathbf{P}$ are losslessly matched, the desired UV mapping relations can be immediately derived: for the $i$-th 3D point $\mathbf{p}_{i} \in \mathbf{P}$, suppose that it is matched with the $j$-th 3D point $\mathbf{\hat{p}}_{j}$ in $\mathbf{\hat{P}}$; note that $\mathbf{\hat{p}}_{j}$ row-wisely corresponds to the $j$-th 2D point $\mathbf{\hat{q}}_{j}$ in $\mathbf{\hat{Q}}$, hence we can derive that the UV coordinate of $\mathbf{p}_{i}$ should be $\mathbf{\hat{q}}_{j}$.

Unfortunately, reconstructing unstructured point clouds with precise point-wise matching is highly challenging, particularly when dealing with large numbers of points. Still, despite the ambiguity of point-wise UV mapping relations acquired from such a 2D-to-3D wrapping procedure, it provides valuable cues indicating how the 3D surface is roughly transformed from the 2D parameter domain. This observation strongly motivates us to incorporate the two opposite mapping directions into a bi-directional learning framework. To build associations between wrapping and unwrapping processes while promoting mapping bijectivity, we further investigate a cycle mapping framework with consistency constraints. Thus, the overall workflow of our global neural surface parameterization pipeline is featured by bi-directional cycle mapping. In what follows, we will detail the corresponding network components, learning mechanisms, and optimization objectives.

\vspace{+0.2cm}
\subsubsection{Bi-directional Cycle Mapping}
\begin{figure*}[!t]
    \centering
    \includegraphics[width=1.0\linewidth]{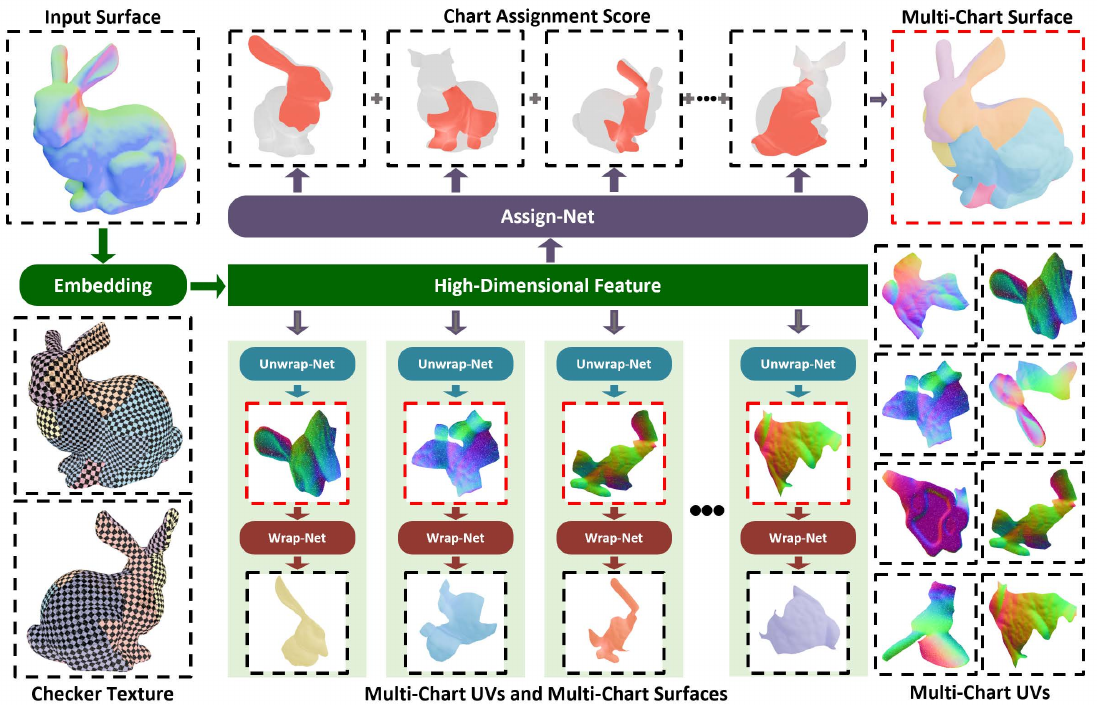}
    \caption{Illustration of multi-chart neural surface parameterization framework, composed of fusion module, chart-assignment module, and multi single directional cycle mapping branches (Unwrap-Net and Wrap-Net). The lower left shows the grid-checkerboard image texture mapping. The right shows an example of a chart assignment result and UV mappings.}
    \label{pipline_multichart}
\end{figure*}
As depicted in Figure \ref{pipline_global}, we build two parallel learning branches consisting of a series of geometrically-interpretable sub-networks built upon point-wisely shared multi-layer perceptrons (MLPs). Here, all sub-networks are parameter-sharing and jointly optimized.

Overall, we develop four types of sub-networks with specific functionalities to mimic the actual physical processes:

\begin{itemize}
	\item The deforming network $\mathcal{M}_{d}(\cdot)$ (Deform-Net) takes as input a set of uniform grid coordinates located at a pre-defined 2D lattice. The initial grids will be adaptively deformed to produce potentially-optimal UV coordinates.
	\item The wrapping network $\mathcal{M}_{w}(\cdot)$ (Wrap-Net) performs 2D-to-3D mapping. Intuitively, the potentially-optimal planar UV points will be smoothly folded to approximate the target 3D surface. 
	\item The cutting network $\mathcal{M}_{c}(\cdot)$ (Cut-Net) operates in the 3D space to derive appropriate cutting seams on the target 3D surface, transforming the original geometric structure into an open and more developable surface manifold. 
	\item The unwrapping network $\mathcal{M}_{u}(\cdot)$ (Unwrap-Net) smoothly flattens 3D surface points onto the 2D parameter domain.
\end{itemize}

We assemble these sub-networks into two parallel learning branches of 2D$\rightarrow$3D$\rightarrow$2D and 3D$\rightarrow$2D$\rightarrow$3D cycle mappings as detailed below.

\vspace{+0.5em}
\noindent\textbf{2D$\rightarrow$3D$\rightarrow$2D Cycle Mapping.} Denote by $\mathbf{G} \in \mathbb{R}^{V\times 2}$ a set of planar grid coordinates uniformly sampled from a pre-defined lattice within $\left[-1, 1 \right]^2$. Treating $\mathbf{G}$ as the initial 2D parameter domain, we tend to explicitly deform it to produce potentially-optimal UV coordinates through the Deform-Net. Concretely, the UV space deformation is implemented as an offset-driven coordinate updating process, which can be formulated as:
\begin{equation}
   \hat{\mathbf{Q}} = \mathcal{M}_d(\mathbf{G}) = \xi_d^{\prime \prime}\left(\left[\xi_d^{\prime}(\mathbf{G}) ; \mathbf{G}\right]\right)+\mathbf{G},
\end{equation}
\noindent where $\mathbf{\hat{Q}} \in \mathbb{R}^{V\times 2}$ is a new set of adaptively-deformed 2D UV coordinates, $\left[ *; * \right]$ denotes channel concatenation, $\xi^{\prime}_{d}(\cdot): \mathbb{R}^{2} \rightarrow \mathbb{R}^{h}$ and $\xi^{\prime\prime}_{d}(\cdot) : \mathbb{R}^{(h+2)} \rightarrow \mathbb{R}^{2}$ are stacked MLP layers. Intuitively, initial 2D grid points are first embedded through $\xi^{\prime}_{d}(\cdot)$ into the $h$-dimensional latent space, then concatenated with itself, and further mapped back onto the planar domain through $\xi^{\prime\prime}_{d}(\cdot)$. The learned offsets are point-wisely added to the initial 2D grid points to produce the resulting $\mathbf{\hat{Q}}$.

After UV space deformation, we feed $\mathbf{\hat{Q}}$ into the Wrap-Net to generate a 3D point cloud $\mathbf{\hat{P}} \in \mathbb{R}^{V\times 3}$, whose underlying surface should approximate the original geometric structure of $\mathbf{P}$. In the meantime, the Wrap-Net is configured with another $3$ output channels to produce point-wise normals $\mathbf{\hat{N}} \in \mathbb{R}^{V\times 3}$. The whole wrapping process can be formulated as:
\begin{equation}
    [\hat{\mathbf{P}} ; \mathbf{\hat{N}}] = \mathcal{M}_w(\hat{\mathbf{Q}}) = \xi_w^{\prime \prime}([\xi_w^{\prime}(\hat{\mathbf{Q}}) ; \hat{\mathbf{Q}})],
\end{equation}
\noindent where $\xi^{\prime}_{w}(\cdot) : \mathbb{R}^{2} \rightarrow \mathbb{R}^{h}$ and $\xi^{\prime\prime}_{w}(\cdot) : \mathbb{R}^{(h+2)} \rightarrow \mathbb{R}^{6}$ are stacked MLP layers. The resulting output channels are respectively split into point-wise coordinates $\mathbf{\hat{P}}$ and normals $\mathbf{\hat{N}}$.

To construct cycle mapping, we sequentially apply the Cut-Net and the Unwrap-Net for the flattening of $\mathbf{\hat{P}}$. Specifically, we transform $\mathbf{\hat{P}}$ to an open 3D manifold $\mathbf{\hat{P}}_\mathrm{cut} \in \mathbb{R}^{V \times 3}$, which should be much more developable than $\mathbf{\hat{P}}$, by learning point-wise offsets:
\begin{equation}
    \hat{\mathbf{P}}_{\mathrm{cut}} = \mathcal{M}_c(\hat{\mathbf{P}}) = \xi_c^{\prime \prime}([\xi_c^{\prime}(\hat{\mathbf{P}}) ; \hat{\mathbf{P}}])+\hat{\mathbf{P}},
\end{equation}
\noindent where $\xi^{\prime}_{c}(\cdot): \mathbb{R}^{3} \rightarrow \mathbb{R}^{h}$ and $\xi^{\prime}_{c^{\prime}}(\cdot): \mathbb{R}^{(h+3)} \rightarrow \mathbb{R}^{3}$ are stacked MLP layers. After that, we perform 3D-to-2D unwrapping as:
\begin{equation}
    \hat{\mathbf{Q}}_{\mathrm{cycle }} = \mathcal{M}_u(\hat{\mathbf{P}}_{\mathrm{cut}}) = \xi_u(\hat{\mathbf{P}}_{\mathrm{cut}}),
\end{equation}
\noindent where $\xi_{u}(\cdot): \mathbb{R}^{3} \rightarrow \mathbb{R}^{2}$ is implemented as stacked MLP layers. The resulting $\mathbf{\hat{Q}}_\mathrm{cycle} \in \mathbb{R}^{V\times 2}$ and the preceding $\mathbf{\hat{Q}}$ form a cycle mapping, since they are supposed to be row-wisely equal.

\vspace{+0.5em}
\noindent\textbf{3D$\rightarrow$2D$\rightarrow$3D Cycle Mapping.} Within the 2D$\rightarrow$3D$\rightarrow$2D cycle mapping branch, the original 3D point set $\mathbf{P}$ just serves as the supervision signal to drive the optimization of $\hat{\mathbf{P}}$, without participation in the actual mapping process. To precisely deduce point-wise UV mappings of $\mathbf{P}$, we further deploy a 3D$\rightarrow$2D$\rightarrow$3D cycle mapping branch, which directly consumes $\mathbf{P}$ at the input end. In this way, when the training process is finished, the networks are adapted to $\mathbf{P}$ to point-wisely produce UV coordinates.

Specifically, we start by feeding the original 3D point set $\mathbf{P}$ into the Cut-Net and the Unwrap-Net to obtain $\mathbf{P}_\mathrm{cut} \in \mathbb{R}^{V\times 3}$ and $\mathbf{Q} \in \mathbb{R}^{V\times 2}$, which can be formulated as:
\begin{equation}
    \mathbf{P}_{\mathrm{cut}}=\mathcal{M}_c(\mathbf{P})=\xi_c^{\prime \prime}\left(\left[\xi_c^{\prime}(\mathbf{P}) ; \mathbf{P}\right]\right)+\mathbf{P},
\end{equation}
\begin{equation}
    \mathbf{Q}=\mathcal{M}_u\left(\mathbf{P}_{\mathrm{cut}}\right)=\xi_u\left(\mathbf{P}_{\mathrm{cut}}\right).
\end{equation}

To construct cycle mapping, we apply the Wrap-Net on the generated 2D UV coordinates $\mathbf{Q}$ to produce $\mathbf{P}_\mathrm{cycle} \in \mathbb{R}^{V\times 3}$ together with point-wise normals $\mathbf{N}_\mathrm{cycle} \in \mathbb{R}^{V\times 3}$, which can be formulated as:
\begin{equation}
    \left[\mathbf{P}_\mathrm{cycle} ; \mathbf{N}_\mathrm{cycle}\right]=\mathcal{M}_w(\mathbf{Q})=\xi_w^{\prime \prime}\left(\left[\xi_w^{\prime}(\mathbf{Q}) ; \mathbf{Q}\right]\right),
\end{equation}
\noindent where $\mathbf{P}_\mathrm{cycle}$ and $\mathbf{N}_\mathrm{cycle} \in \mathbb{R}^{V\times 3}$ form cycle mappings with $\mathbf{P}$ and $\mathbf{N}$, respectively. Thus, $\mathbf{P}_\mathrm{cycle}$ should be row-wisely equal to $\mathbf{P}$, the directions of normal vectors $\mathbf{N}_\mathrm{cycle}$ and $\mathbf{N}$ should also be row-wisely consistent.

Particularly, it is worth noting that, throughout the overall bi-directional cycle mapping workflow, \textbf{\textit{only $\mathbf{Q}$ is regarded as the desired 2D UV coordinates of the original 3D points $\mathbf{P}$}}, i.e., $\mathbf{P}$ and $\mathbf{Q}$ are row-wisely corresponded. All the other intermediate variables, such as $\mathbf{\hat{Q}}$ and $\mathbf{\hat{Q}}_\mathrm{cycle}$, are merely used to facilitate the overall optimization process. This is because the correspondences of either $\mathbf{\hat{Q}}$ or $\mathbf{\hat{Q}}_\mathrm{cycle}$ between $\mathbf{P}$ remain unknown, while $\mathbf{Q}$ is exactly point-wisely mapped from $\mathbf{P}$.

Moreover, it is possible to explicitly extract cutting seams that are implicitly learned through the Cut-Net. To identify a set of 3D surface points $\mathbf{C} = \{\mathbf{c}_i \in \mathbf{P} \}$ located on the cutting seams, we directly compare the mapping relationship between $\mathbf{P}$ and $\mathbf{Q}$. Specifically, for the $i$-th 3D point ${\mathbf{p}_{i} \in \mathbf{P}}$ together with its neighboring 3D points $\{\mathbf{p}_{i, j}\}^{J_\mathrm{cut}}_{j=1}$ within $\mathbf{P}$, their UV coordinates are $\mathbf{q}_{i} \in \mathbf{Q}$ and $\{ \mathbf{q}_{i, j} \}^{J_\mathrm{cut}}_{j=1}$. The maximum distance $\eta_{i}$ between $\mathbf{q}_i$ and its neighbors can be computed as:
\begin{equation} \label{eqn:find-cut}
	\eta_i = \mathrm{max}(\{ \lVert \mathbf{q}_i - \mathbf{q}_{i, j} \rVert_2 \}_{j=1}^{J_\mathrm{cut}}).
\end{equation}

If $\eta_{i}$ exceeds a threshold $\tau$, we consider $\mathbf{p}_i$ to locate on the underlying cutting seams. Examining all 3D points within $\mathbf{P}$ can produce a subset of seam points $\mathbf{C}$. In this way, no seam points would be identified if cutting the original 3D surface is actually unnecessary.

\vspace{+0.2cm}
\subsubsection{Training Objectives}

As an unsupervised learning architecture, we need to optimize our bi-directional cycle mapping process with carefully designed objective/loss functions and constraints, detailed as follows.

\vspace{+0.5em}
\noindent\textbf{Unwrapping Loss.} The most fundamental requirement for the generated UV coordinates $\mathbf{Q}$ is to avoid mutual overlappings. For $\mathbf{q}_{i} \in \mathbf{Q}$ and its neighboring points $\{ \mathbf{q}_{i, j} \}^{J_\mathrm{u}}_{j=1}$, we penalize pairs of neighbors that are too close:
\begin{equation}\label{eqn:unwapping-loss}
    \ell_u=\sum_{i=1}^V \sum_{j=1}^{J_{\mathrm{u}}} \max (0, \epsilon-\left\|\mathbf{q}_i-\mathbf{q}_{i, j}\right\|_2),
\end{equation}
\noindent where the minimum distance between each pair of neighboring points within $\mathbf{Q}$ is constrained by a threshold $\epsilon$.  

\vspace{+0.5em}
\noindent\textbf{Wrapping Loss.}  For 2D$\rightarrow$3D$\rightarrow$2D cycle mapping, the major supervision is that the underlying 3D surface of the generated point set $\mathbf{\hat{P}}$ is supposed to approximate the original  geometric structure. Note that the correspondence between $\mathbf{\hat{P}}$ and $\mathbf{P}$ is unknown. Hence, we minimize the Chamfer Distance, denoted as $\operatorname{CD}(*; *)$, to measure the similarity between point sets:
\begin{equation}
    \ell_w=\operatorname{CD}(\hat{\mathbf{P}} ; \mathbf{P}).
\end{equation}

\vspace{+0.5em}
\noindent\textbf{Cycle Consistency Loss.} The symmetric designs of the sub-networks within our bi-directional cycle mapping framework facilitate promoting point-wise consistencies between $\mathbf{P}$ and $\mathbf{P}_{\mathrm{cycle}}$, $\mathbf{N}$ and $\mathbf{N}_{\mathrm{cycle}}$, as well as $\hat{\mathbf{Q}}$ and $\hat{\mathbf{Q}}_{\mathrm{cycle}}$, by minimizing:
\begin{equation}\label{eqn:cycle-consistency-loss}
    \ell_c = \left\|\mathbf{P}-\mathbf{P}_{\mathrm{cycle }}\right\|_{1}+\|\hat{\mathbf{Q}}-\hat{\mathbf{Q}}_{\mathrm{cycle }}\|_{1}+\operatorname{CS}(\mathbf{N}; \mathbf{N}_{\mathrm{cycle }}),
\end{equation}
where $\operatorname{CS}(*; *)$ computes point-wise cosine similarity between normal vectors.

\vspace{+0.5em}
\noindent\textbf{Distortion Constraint.} The ideal parameterization is supposed to achieve zero distortion under isometric measurement, which is basically impossible in real applications. In our implementation for global parameterization, we constrain the conformal distortion to promote angle-preserving mappings.

We adopt a hybrid distortion computation scheme composed of a differential distortion loss (DDL) and a triangle distortion loss (TDL). The DDL implicitly constrains the neural functions, while the TDL explicitly measures pairs of 3D and 2D triangles. The combination of the two forms of loss functions can effectively achieve low-distortion parameterization.

Specifically, the DDL regularizes neural layers by exploiting differential surface properties that can be conveniently deduced via the automatic differentiation mechanisms in common deep learning programming frameworks. During bi-directional cycle mapping, $\mathbf{Q}$ is mapped to $\mathbf{P}_{\mathrm{cycle}}$, and $\mathbf{\hat{Q}}$ is mapped to $\mathbf{\hat{P}}$. We denote such two mappings as $\varphi: \mathbb{R}^{2}\rightarrow\mathbb{R}^{3}$ and $\hat{\varphi}: \mathbb{R}^{2}\rightarrow\mathbb{R}^{3}$. Then we compute the derivatives of $\varphi$ and $\hat{\varphi}$ for each 2D UV point coordinate, producing two Jacobian matrices $\mathbf{B} \in \mathbb{R}^{3 \times 2}$ and $\mathbf{\hat{B}} \in \mathbb{R}^{3 \times 2}$ as:
\begin{equation}
    \mathbf{B} = (\varphi_{u}\ \varphi_{v}),\; \mathbf{\hat{B}} = (\hat{\varphi}_{u}\ \hat{\varphi}_{v}),
\end{equation}
\noindent where $\varphi_{u}$, $\varphi_{v}$, $\hat{\varphi}_{u}$, and $\hat{\varphi}_{v}$ are 3D vectors of partial derivatives. After that, we compute the eigenvalues of $\mathbf{B}^T \mathbf{B}$ and $\mathbf{\hat{B}}^T \mathbf{\hat{B}}$, which are denoted as $(\lambda_{1}, \lambda_{2})$ and $(\hat{\lambda}_{1}, \hat{\lambda}_{2})$. Thus, we formulate the DDL as:
\begin{equation}
    \ell_{\mathrm{diff}} = \sum_{\mathbf{q}\in \mathbf{Q}}\left\|\lambda_1-\lambda_2\right\|_{1}+\sum_{\mathbf{\hat{q}}\in \mathbf{\hat{Q}}}\left\|\hat{\lambda}_{1}-\hat{\lambda}_{2}\right\|_{1}.
\end{equation}

For the computation of TDL, we measure the angle differences within each pair of 3D and 2D triangles. Given a certain triangle $\Omega_\mathbf{P}$ in the original mesh $\mathcal{X}$ with three angles $\{ \theta_i \}_{i=1}^3$, since each triangle vertex in $\mathbf{P}$ is mapped to its UV coordinate in $\mathbf{Q}$, we can obtain the corresponding 2D triangle $\Omega_\mathbf{Q}$ within the UV space with three corresponding angles $\{ \beta_i \}_{i=1}^3$. Thus, we formulate the TDL as:
\begin{equation}
    \ell_{\mathrm{tri}} = \sum\nolimits_{\Omega_\mathbf{P} \in \mathcal{X}} \sum\nolimits_{i}\left\|\theta_i - \beta_i\right\|_{1}.
\end{equation}

The overall loss function for global parameterization can be formulated as:
\begin{equation}\label{eqn:global-loss}
	\ell_\mathrm{global} = \alpha_{u} \cdot \ell_u + \alpha_{w} \cdot \ell_w + \alpha_{c} \cdot \ell_c + \alpha_\mathrm{diff} \cdot \ell_{\mathrm{diff}} + \alpha_\mathrm{tri} \cdot \ell_{\mathrm{tri}},
\end{equation}
where the weights $\alpha_{u}$, $\alpha_{w}$, $\alpha_{c}$, $\alpha_\mathrm{diff}$ and $\alpha_\mathrm{tri}$ are empirically adjusted to balance the effects of different terms.

\subsection{Multi-Chart Neural Surface Parameterization} \label{sec:method-local-para}

Achieving multi-chart parameterization requires decomposing the complete geometric surface into multiple segments, each of which should be easier to flatten with reduced distortion.

Given a pre-specified number of charts $K$, we tend to generate multiple sets of 2D UV coordinates $\{ \mathbf{L}^{(k)} \in \mathbb{R}^{V_k \times 2} \}_{k=1}^K$ correspondingly mapped from $\{ \mathbf{P}^{(k)} \in \mathbb{R}^{V_k \times 3} \}_{k=1}^K$, such that $\sum\nolimits_k V_k = V$ and each $\mathbf{P}^{(k)}$ is exactly a subset of $\mathbf{P}$.

\vspace{+0.2cm}
\subsubsection{Architecture Adaptations}

We implement the multi-chart parameterization framework by adapting our preceding global parameterization framework with necessary modifications and particularly customized network structures and loss functions.

As illustrated in Figure \ref{pipline_multichart}, we introduce a chart assignment sub-network $\mathcal{M}_a(\cdot)$ (Assign-Net) to explicitly measure the probability of each 3D point belonging to all K charts. Additionally, different from the preceding bi-directional learning mechanism of our global parameterization framework, we only employ the 3D$\rightarrow$2D$\rightarrow$3D cycle mapping for multi-chart parameterization, which is composed of more light-weight versions of Unwrap-Net and Wrap-Net components. Since local surface charts are typically much easier to flatten, such architectural simplification can significantly improve efficiency while maintaining parameterization performance.

Technically, we start by embedding the original point set $\mathbf{P}$ into the latent space to produce $d$-dimensional feature vectors $\mathbf{H} \in \mathbb{R}^{V \times d}$. To achieve adaptively-learnable chart assignment, we apply the Assign-Net to explicitly generate $K$ assignment scores for each 3D point, which can be formulated as:
\begin{equation}
	\mathbf{S} = \mathcal{M}_a(\mathbf{H}) = \operatorname{Softmax}(\xi_{a}(\mathbf{H})),
\end{equation}
\noindent where $\mathbf{S} \in \mathbb{R}^{V \times K}$ encodes point-wise chart assignment scores, $\xi_{a}(\cdot): \mathbb{R}^{d} \rightarrow \mathbb{R}^{K}$ is implemented as stacked MLP layers, and we apply column-wise Softmax to deduce the probability distribution.

In the meantime, we concurrently feed $\mathbf{H}$ into the $K$ sets of independent (non-weight-sharing) Unwrap-Net and Wrap-Net components to perform cycle mappings in parallel, where the input channels of these Unwrap-Nets are accordingly adjusted to $d$. The overall mapping process can be described by:
\begin{equation}
    \{ \mathbf{Q}^{(k)} \}_{k=1}^K = \{ \mathcal{M}_u^{(k)}(\mathbf{H}) \}_{k=1}^K,
\end{equation}
\begin{equation}
    \{ [\mathbf{P}_{\mathrm{cycle}}^{(k)} ; \mathbf{N}_{\mathrm{cycle}}^{(k)}] \}_{k=1}^K = \{ \mathcal{M}_w^{(k)}(\mathbf{Q}^{(k)}) \}_{k=1}^K,
\end{equation}
\noindent where $\mathbf{Q}^{(k)} \in \mathbb{R}^{V \times 2}$, $\mathbf{P}_{\mathrm{cycle}}^{(k)} \in \mathbb{R}^{V \times 3}$, and $\mathbf{N}_{\mathrm{cycle}}^{(k)} \in \mathbb{R}^{V \times 3}$.

It is worth noting that the resulting $\{ \mathbf{Q}^{(k)} \}_{k=1}^K$ are different from our actually desired $\{ \mathbf{L}^{(k)} \}_{k=1}^K$. Below, we introduce our customized supervision mechanisms to drive the optimization of chart-wise surface parameterization.

\vspace{+0.2cm}
\subsubsection{Chart-wise Score-Weighted Loss Functions}

We employ a score-weighted supervision mechanism to drive the adaptive chart assignment as well as chart-wise parameterization. This is achieved by multiplying the different loss terms point-wisely with explicitly learned chart assignment scores, such that each 3D$\rightarrow$2D$\rightarrow$3D mapping cycle exclusively focuses on minimizing the parameterization distortion of its corresponding chart. Meanwhile, the Assign-Net is also naturally driven to update the output chart assignment scores, such that each surface chart is as developable as possible.

For convenience, the column vectors in $\mathbf{S}$ are split as $\mathbf{S} = \{ \mathbf{s}^{(1)}, \cdots, \mathbf{s}^{(k)}, \cdots, \mathbf{s}^{(K)} \}$ to separately represent the point-wise scores for the assignment of the $k$-th chart, where $\mathbf{s}^{(k)} \in \mathbb{R}^V$ represents the probability of each vertex being assigned to the $k$-th chart.

To avoid overlapped UV coordinates, the unwrapping loss for each chart is formulated as:
\begin{equation}
    \ell_u^{(k)}=\sum_{i=1}^N \sum_{k=1}^{K_{\mathrm{u}}} \max \left(0, \epsilon-\|(\mathbf{q}_i-\mathbf{q}_i^{(k)}) \cdot \mathbf{s}^{(k)} \|_2\right).
\end{equation}

Within the 3D$\rightarrow$2D$\rightarrow$3D mapping process, we can promote cycle consistency as:
\begin{equation}
    \ell_c^{(k)} = \big\|(\mathbf{P}-\mathbf{P}_{\mathrm{cycle}}^{(k)}) \cdot \mathbf{s}^{(k)} \big\|_{1} + \operatorname{CS}(\mathbf{N}; \mathbf{N}_{\mathrm{cycle}}^{(k)}; \mathbf{s}^{(k)}),
\end{equation}
\noindent where $\operatorname{CS}(\mathbf{N}; \mathbf{N}_{\mathrm{cycle }}^{(k)}; \mathbf{s}^{(k)})$ means point-wisely multiplying the chart assignment score $\mathbf{s}^{(k)}$ with the cosine distances between $\mathbf{N}$ and $\mathbf{N}_{\mathrm{cycle}}^{(k)}$, such that each of the $K$ mapping cycles merely focuses on the point-wise consistency within its own chart.

For the distortion constraint, we only compute the TDL term chart-wisely. Besides, considering that local surface charts are easier to flatten, here we impose stronger isometric constraints.
\begin{figure*}[!t]
    \centering
    \includegraphics[width=1\linewidth]{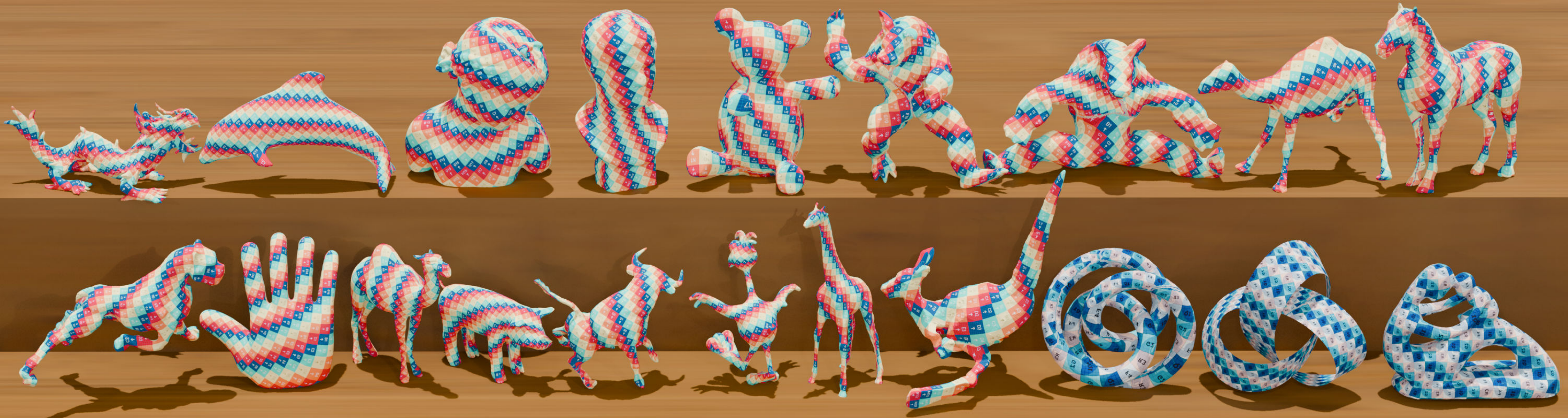}
    \caption{A gallery for our global parameterization results, utilizing a grid-checkboard image to illustrate the parameterization deformation. Standard models (with the red-blue gradient grid-checkerboard image) and complex topology models (with the blue grid-checkerboard image) are demonstrated.}
    \label{gallery_global}
\end{figure*}
\begin{figure*}[!t]
    \centering
    \includegraphics[width=1\linewidth]{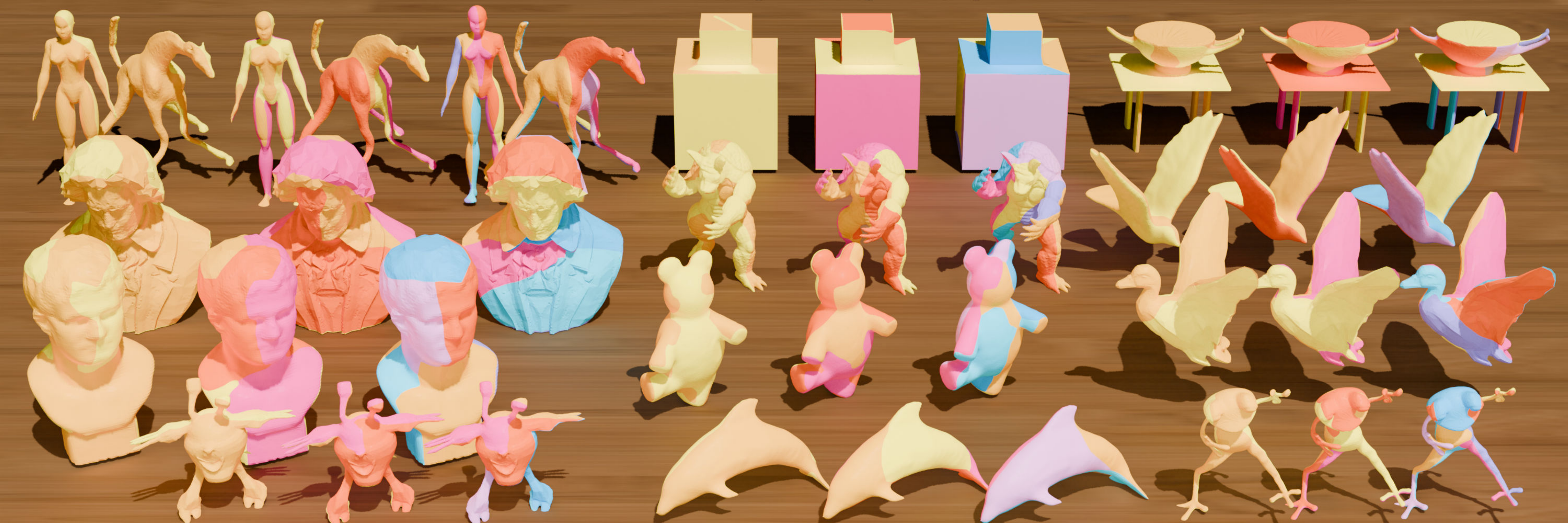}
    \caption{A gallery for our multi-chart parameterization results. It shows segmentation results for configurations utilizing 2, 4, and 8 charts (from left to right).}
    \label{gallery_multichart}
\end{figure*}
For an edge $\mathbf{e}$ belonging to a certain 3D triangle with two end points $\mathbf{p}_i$ and $\mathbf{p}_j$, whose chart assignment scores with respect to the $k$-th chart are $\mathbf{S}(i, k)$ and $\mathbf{S}(j, k)$, we deduce a chart assignment score for this edge by averaging the scores of the two end points, as given by:
\begin{equation}
	\mathbf{s}^{(k)}(\mathbf{e}) = (\mathbf{S}(i, k) + \mathbf{S}(j, k)) / 2.
\end{equation}

Let $\mathbf{\tilde{e}}^{(k)}$ be the correspondingly mapped 2D triangle edge with end points $\mathbf{q}^{(k)}_i$ and $\mathbf{q}^{(k)}_j$. The distortion constraint of this particular triangle pair is computed as:
\begin{equation}
	\ell_\mathrm{edge}^{(k)}(\mathbf{e}) = \mathbf{s}^{(k)}(\mathbf{e}) \cdot \big\| (\big\| \mathbf{p}_i - \mathbf{p}_j \big\|_2 - \big\| \mathbf{q}^{(k)}_i - \mathbf{q}^{(k)}_j \big\|_2) \big\|_2^2.
\end{equation}
Thus, the overall distortion constraint is computed by iterating all pairs of triangle edges:
\begin{equation}
	\ell_\mathrm{tri}^{(k)} = \sum\nolimits_{\Omega_\mathbf{P} \in \mathcal{X}} \sum\nolimits_{\mathbf{e} \in \Omega_\mathbf{P}} \ell_\mathrm{edge}^{(k)}(\mathbf{e}).
\end{equation}

The overall loss function of all charts is formulated as:
\begin{equation}
	\ell_\mathrm{local} = \sum_{k=1}^K   \alpha_{u} \cdot {\ell_u^{(k)}} + \alpha_{c} \cdot {\ell_c^{(k)}} + \alpha_{\mathrm{tri}} \cdot {\ell_\mathrm{tri}^{(k)}},
\end{equation}
\noindent where $\alpha_{u}$, $\alpha_{c}$ and $\alpha_{\mathrm{tri}}$ are weights for different loss terms.

Finally, we can obtain the desired $K$ sets of $\mathbf{L}^{(k)}$ and $\mathbf{P}^{(k)}$ from the generated $\mathbf{Q}^{(k)}$ and the original $\mathbf{P}$ based on the chart assignment score matrix $\mathbf{S}$, which can be formulated as:
\begin{equation}
    \mathbf{L}^{(k)} = \{\mathbf{q}_{i}^{(k)}\in \mathbf{Q}^{(k)}|\arg\max_j \mathbf{S}(i,j)=k \},
\end{equation}
\begin{equation}
    \mathbf{P}^{(k)} = \{\mathbf{p}_{i} \in \mathbf{P} | \arg\max_j \mathbf{S}(i,j) = k \}.
\end{equation}

\section{Experiments} \label{Experiment}

We evaluated our neural surface parameterization approach both qualitatively and quantitatively. First, we presented galleries of our global and multi-chart parameterization methods to intuitively demonstrate their effectiveness. Second, we compared our methods with commonly used approaches for global and multi-chart parameterization tasks, providing both visual and quantitative comparisons. Finally, we conducted ablation studies to validate the effectiveness and necessity of our architectural design.
\begin{figure*}[!t]
    \centering
    \includegraphics[width=1.0\linewidth]{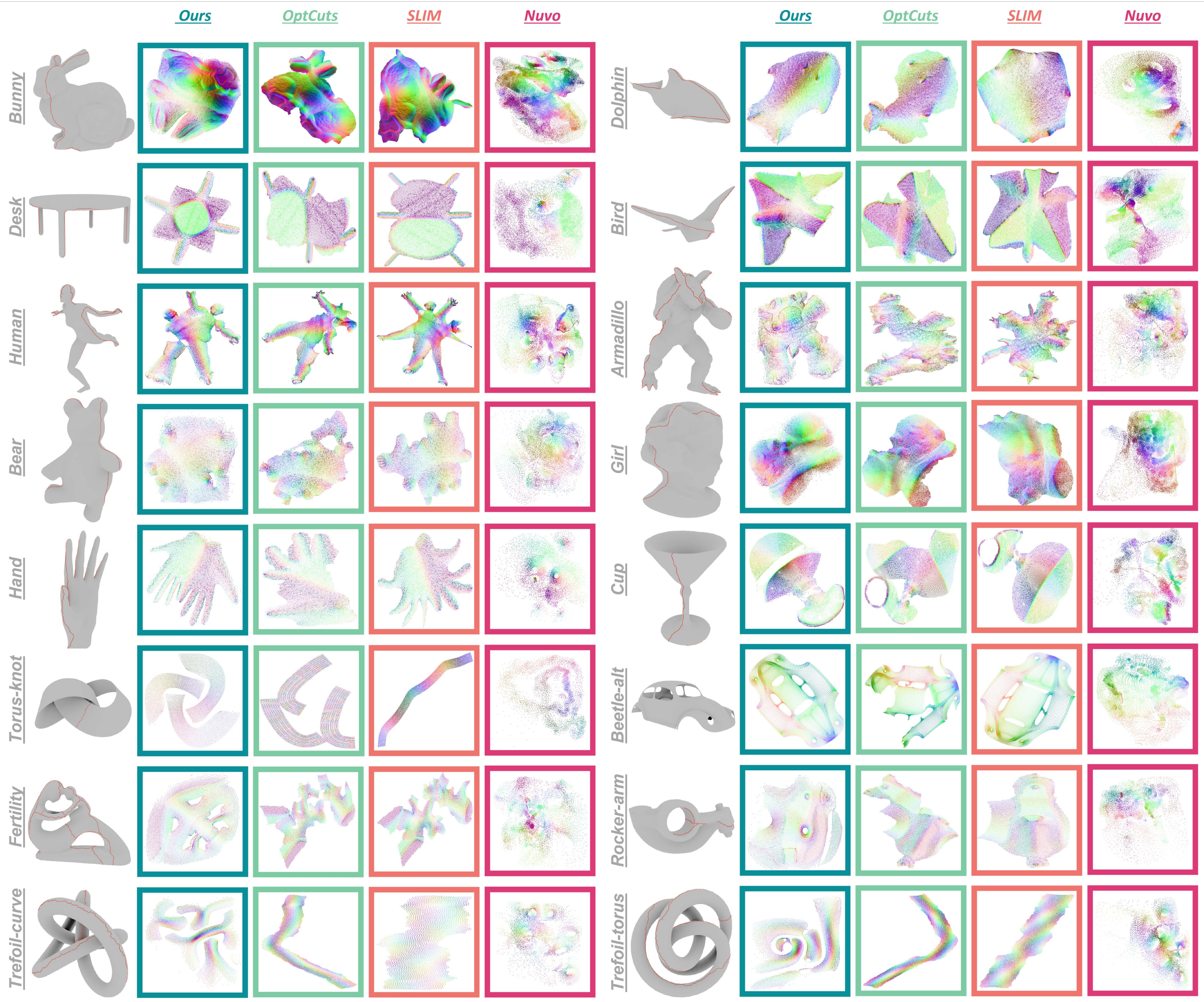}
    
    \caption{\revise{Comparison of UV unwrapping results on different surface models produced by our global neural surface parameterization method, OptCuts, SLIM, and Nuvo ($n=1$), where the 2D UV coordinates are color-coded by ground-truth point-wise normals to facilitate visualization. Red lines marked on the model represent the seams inputs required for the SLIM.}}
    \label{visual_comparison_global}
\end{figure*}

\subsection{Implementation Details}
In our global neural surface parameterization framework, all sub-networks are architecturally built upon stacked MLPs without batch normalization. We uniformly configured LeakyReLU non-linearities with the negative slope of 0.01, except for the output layer. Within the Deform-Net, $\xi_d(\
\cdot)$ and $\xi_d^{\prime}(\cdot)$ are four-layer MLPs with channels [2, 512, 512, 512, 64] and [66, 512, 512, 512, 2]. Within the Wrap-Net, $\xi_w(\cdot)$ and $\xi_w^{\prime}(\cdot)$ are four-layer MLPs with channels [2, 512, 512, 512, 64] and [66, 512, 512, 512, 6]. Within the Cut-Net, $\xi_c(\cdot)$ and $\xi_c^{\prime}(\cdot)$ are three-layer MLPs with channels [3, 512, 512, 64] and [67, 512, 512, 3]. Within the Unwrap-Net, $\xi_u(\cdot)$ is implemented as three-layer MLPs with channels [3, 512, 512, 2]. In addition to network structures, there is also a set of hyperparameters to be configured and tuned. As presented in Eq. \eqref{eqn:find-cut} for cutting seam extraction, we choose $J_{\mathrm{cut}} = 3$. Suppose that, at the current training iteration, the side length of the square bounding box of 2D UV coordinates $\mathbf{Q}$ is denoted as $L(\mathbf{Q})$. Then we set the threshold $\tau$ to be 2\% of $L(\mathbf{Q})$. Besides, as presented in Eq. \eqref{eqn:unwapping-loss} for computing the unwrapping loss, we choose the threshold as $\epsilon = 0.2 \cdot L(\mathbf{Q})/\sqrt{N}$. When formulating the overall training objective, the weights $\alpha_{u}$, $\alpha_{w}$, $\alpha_{c}$, $\alpha_{\mathrm{diff}}$, and $\alpha_{\mathrm{tri}}$ are set as 0.01, 1.0, 0.01, 0.01 and 0.001, resepctively. 

In our multi-chart neural surface parameterization framework, the settings are basically the same as the global parameterization. The embedding module is a  two-layer MLPs with channels $[3, 512, 512]$. Within the Assign-Net, $\xi_{a}(\cdot)$ is a two-layer MLPs with channels $[512, 512, K]$. Within the Unwrap-Net, the channels are changed to $[512,512,512,2]$ with a high-dimensional feature vector as input. When formulating the overall training objective, the weights $\alpha_{u}$, $\alpha_{c}$, and $\alpha_{\mathrm{tri}}$ are set as 0.01, 10 and 1. 

All our experiments were conducted on a single NVIDIA GeForce RTX 3090 GPU.

\subsection{Gallery}
We present two galleries to demonstrate the effectiveness of our method intuitively: one for global parameterization and another for multi-chart parameterization. For the global parameterization gallery, we selected a series of representative models with intricate geometric structures, complex topologies, and high genus. To quantify distortion in the parameterization process, we used grid-checkboard images. As shown in Figure \ref{gallery_global}, our parameterization results exhibit remarkably low distortion and consistently perform well across models with intricate geometric details, complex topologies, and high genus.

\begin{table}[!t]
    \caption{Global parameterization quantitative comparisons of our method, OptCuts \cite{li2018optcuts}, SLIM \cite{rabinovich2017scalable} and Nuvo \cite{srinivasan2025nuvo} in terms of parameterization conformity. The best results are highlighted in \textbf{bold}.} \vspace{-0.3cm}
    \centering
    \renewcommand{\arraystretch}{1.5}
    \begin{tabular}{c|c|c|c|c}
    \toprule[1.5pt]
    Model & Ours & OptCuts  & SLIM & Nuvo \\
    \hline
    Bunny & \textbf{0.0634} & 0.0714 & 0.0939 & 0.6912\\
    \hline
    Dolphin & \textbf{0.0282} &0.0799 &0.3360 &0.7584\\
    \hline
    Desk & \textbf{0.0332} &0.0730 &0.0396 &0.6843\\
    \hline
    Bird &\textbf{0.0371} &0.0756 &0.0795 &0.7127\\
    \hline
    Human & \textbf{0.0835} &0.0986 &0.0957 &0.6681\\
    \hline
    Armadillo & \textbf{0.0779} &0.0963 &0.1061 &0.6574\\
    \hline
    Bear & \textbf{0.0457} &0.0912 &0.0727 &0.6968\\
    \hline
    Gril &\textbf{0.0462} &0.0807 &0.0769 &0.6752\\
    \hline
    Hand & \textbf{0.0359} &0.0836 &0.0477 &0.7141\\
    \hline
    Cup & \textbf{0.0508} &0.0911 &0.1732 &0.6782\\
    \hline
    Torus-knot & \textbf{0.0150} &0.7320 &0.0439 &1.0312\\
    \hline
    Beetle-alt & \textbf{0.0128} &0.0213 &0.0364 &0.2974\\
    \hline
    Fertility & \textbf{0.0538} &0.1152 &0.1092 &1.1560\\
    \hline
    Rocker-arm & \textbf{0.0641} &0.0963 &0.1764 &0.4463\\
    \hline
    Trefoil-curve & \textbf{0.0470} &0.1014 &0.0684 &0.8556\\
    \hline
    Trefoil-torus & \textbf{0.0564} &0.0930 &0.0641 &0.8251\\
    \bottomrule[1.5pt]
    \end{tabular}
    \label{quantitative_comparison_global}
\end{table}
\begin{table}[!t]
    \caption{Global parameterization quantitative comparisons of our method, OptCuts \cite{li2018optcuts}, SLIM \cite{rabinovich2017scalable} and Nuvo \cite{srinivasan2025nuvo} in terms of optimization time (seconds). The best results are highlighted in \textbf{bold}.} \vspace{-0.3cm}
    \centering
    \renewcommand{\arraystretch}{1.5}
    \begin{tabular}{c|c|c|c|c}
    \toprule[1.5pt]
    Model & Ours & OptCuts  & SLIM & Nuvo \\
    \hline
    Bunny & 405&	1286&	\textbf{77}&	539\\
    \hline
    Dolphin &397	&293	&\textbf{22}	&554\\
    \hline
    Desk & 389	&179	&\textbf{14}	&541\\
    \hline
    Bird &411	&387	&\textbf{26}&	538\\
    \hline
    Human & 412	&742	&\textbf{18}	&512\\
    \hline
    Armadillo & 403	&319&	\textbf{13}	&541\\
    \hline
    Bear & 407	&85	&\textbf{9}	&556\\
    \hline
    Gril &418	&287	&\textbf{30}&	560\\
    \hline
    Hand & 405	&98	&\textbf{6}	&543\\
    \hline
    Cup & 380	&214	&\textbf{15}	&561\\
    \hline
    Torus-knot & 391	&\textbf{2}	&\textbf{2}	&533\\
    \hline
    Beetle-alt & 385	&9	&\textbf{3}	&554\\
    \hline
    Fertility & 397	&59	&\textbf{5}	&531\\
    \hline
    Rocker-arm &399&	22	&\textbf{4}	&548\\
    \hline
    Trefoil-curve & 413	&4	&\textbf{3}	&527\\
    \hline
    Trefoil-torus &398	&6	&\textbf{2}	&529\\
    \bottomrule[1.5pt]
    \end{tabular}
    \label{quantitative_comparison_global_time}
\end{table}
In the context of multi-chart parameterization, we present segmentation results for configurations using 2, 4, and 8 charts. As shown in Figure \ref{gallery_multichart}, our multi-chart parameterization framework generates geometrically meaningful segmentations, driven by the goal of achieving low-distortion UV mappings. The proposed method consistently produces effective segmentation outcomes across diverse models, including those with smooth surfaces and those with sharp edges. Increasing the number of charts generally enhances the ability to achieve parameterization with lower distortion. However, our approach prioritizes distortion minimization while avoiding unnecessary segmentation. Consequently, when a higher number of charts is specified, the algorithm adaptively uses fewer charts if they suffice to maintain low distortion levels. This behavior is evident in Figure \ref{gallery_multichart}, where, for a specified value of $n=8$, certain models are parameterized with fewer than eight charts. This outcome suggests that, for these models, additional charts do not significantly further reduce distortion.

\begin{figure}[!t]
    \centering
    \includegraphics[width=1.0\linewidth]{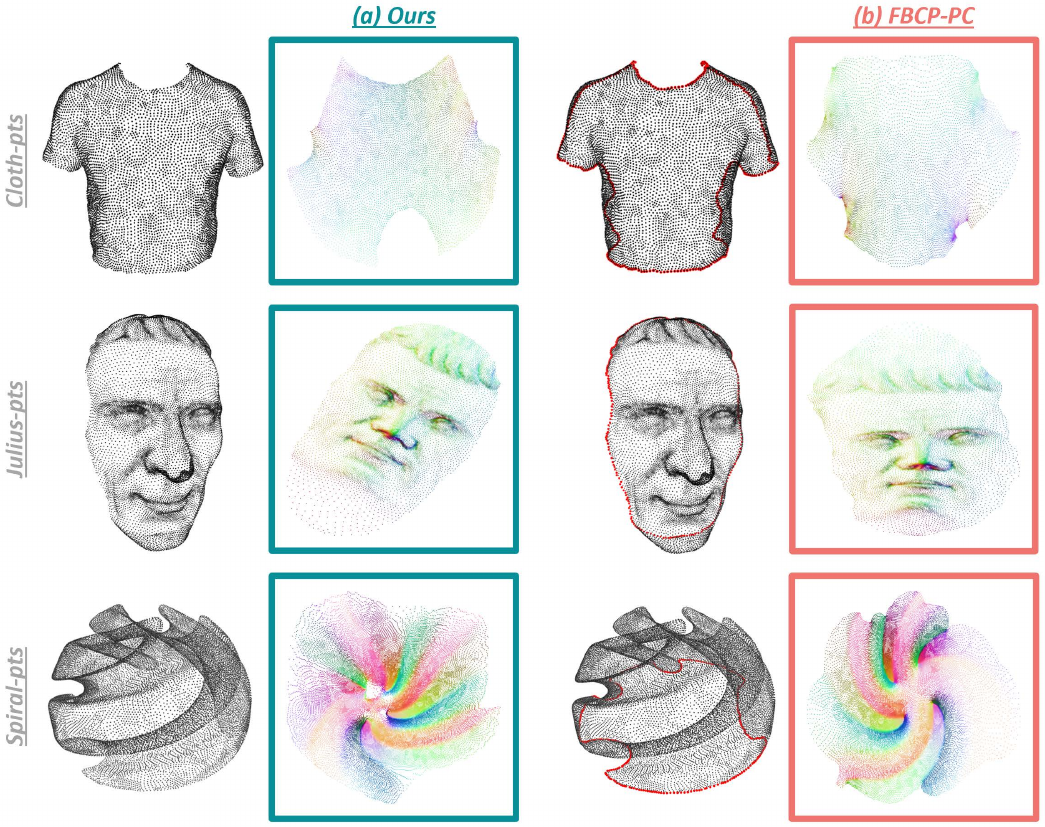}
    \caption{Point cloud parameterization achieved by (a) our method and (b) FBCP-PC.}
    \label{point_cloud_parameterization}
\end{figure}

\subsection{Comparison on Global Parameterization}
To evaluate the performance of global parameterization, we selected a diverse set of models with varying geometric structures, topologies, and genera. For comparative analysis, we employed three methods: OptCuts \cite{li2018optcuts}, SLIM \cite{rabinovich2017scalable}, and Nuvo \cite{srinivasan2025nuvo}. Nuvo, similar to our proposed approach, is a neural radiance field-based parameterization technique that supports both global and multi-chart parameterizations. In contrast, OptCuts is designed to optimize seams and parameterization mappings simultaneously. Most parameterization methods that map 3D surfaces onto 2D planes rely on high-quality seam inputs and lack the ability to jointly optimize seams and mappings. As a representative of such methods, we selected SLIM and tested it using \textbf{manually provided} high-quality seams. \revise{Manually specified secants are typically placed along the relatively sharp edges of the model's geometry to facilitate the parameterization method in achieving UVs with less distortion.}
\begin{table}[!t]
    \caption{Global parameterization quantitative comparisons of our method and FBCP-PC \cite{choi2022free} in terms of parameterization conformity. The best results are highlighted in \textbf{bold}.} \vspace{-0.3cm}
    \centering
    \renewcommand{\arraystretch}{1.5}
    \begin{tabular}{c|c|c}
    \toprule[1.5pt]
    Model & Ours &  FBCP-PC \\
    \hline
    Cloth-pts (\#Pts=7K) & 0.022 & \textbf{0.021}\\
    \hline
    Julius-pts (\#Pts=11K)& 0.058 &\textbf{0.019}\\
    \hline
    Spiral-pts (\#Pts=28K)& 0.098 &\textbf{0.023}\\
    \bottomrule[1.5pt]
    \end{tabular}
    \label{quantitative_point_cloud_parameterization}
\end{table}

Our evaluation focuses on two key aspects: visual quality and conformity preservation. To this end, we used all mesh vertices, computing their corresponding positions in UV space. The parameterization results for all four methods are illustrated in Figure \ref{visual_comparison_global}, where surface normals are used to color the visualizations. For quantitative comparison, we calculated conformality metrics—lower values indicate better performance—by measuring the average absolute angular difference between corresponding triangles in the 3D mesh and their 2D parameterized representations. These results are summarized in Table \ref{quantitative_comparison_global}.

Optimizing seams for parameterization is an inherently complex task. Consequently, comparing our method to SLIM, which relies on manually specified high-quality seams, introduces an element of unfairness. When provided with optimal seams, SLIM achieves excellent parameterization results. Nevertheless, as shown in Figure \ref{visual_comparison_global}, our method produces UV unwrapping with comparable visual quality. Furthermore, in cases where seam quality is suboptimal, our approach outperforms SLIM. This advantage is supported by the conformality metrics in Table \ref{quantitative_comparison_global}, which show that our method consistently achieves the lowest conformal error across most tested models. 

Table \ref{quantitative_comparison_global_time} shows a comparison of optimization times. SLIM achieves the lowest optimization time by leveraging pre-provided seams, thus avoiding seam optimization. The efficiency of OptCuts varies significantly across different models: it performs poorly when the ideal seam is complex, but approaches the efficiency of SLIM when the ideal secant is simpler. As neural network-based approaches, our method outperforms Nuvo in efficiency, as Nuvo relies on neural radiance fields, requiring more extensive training on densely sampled points of models.

Our global parameterization method achieves superior performance on mesh surfaces with connectivity. Additionally, our approach is equally applicable to the parameterization of point clouds without connectivity. To accommodate unstructured and unoriented point clouds, our framework requires only the removal of normal-driven components, i.e., the last term $\operatorname{CS}(\mathbf{N}; \mathbf{N}_{\mathrm{cycle }})$ in Eq. \eqref{eqn:cycle-consistency-loss} and TDL term $\alpha_\mathrm{tri}\cdot \ell_\mathrm{tri}$ in Eq. \eqref{eqn:global-loss}. As compared in Figure \ref{point_cloud_parameterization} and Table \ref{quantitative_point_cloud_parameterization}, our performances are not better than FBCP-PC. However, we must point out that FBCP-PC \cite{choi2022free} requires \textbf{manually specifying} indices of boundary points (arranged in order) as additional inputs, which is a strong priori. Hence, the comparisons are actually quite unfair to us.

\begin{figure*}[!t]
    \centering
    \includegraphics[width=0.99\linewidth]{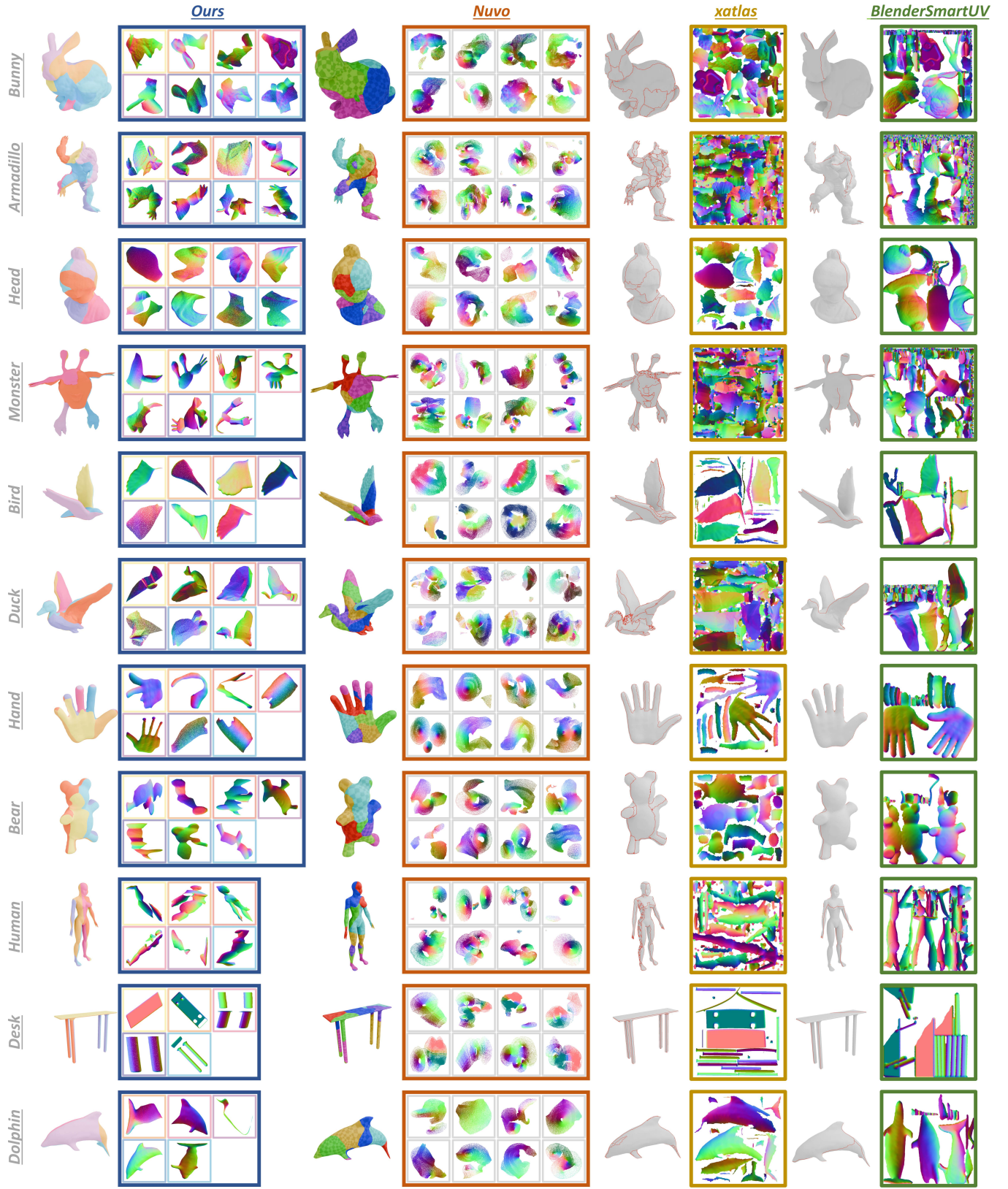}
    \vspace{-0.1cm}
    \caption{Comparison of multi-chart UV unwrapping results on different surface models produced by our multi-chart neural surface parameterization method, Nuvo, xatlas and Blender Smart UV, where the 2D UV coordinates are color-coded by ground-truth point-wise normals to facilitate visualization.}
    \label{visual_comparison_multi_chart}
    \vspace{-0.5cm}
\end{figure*}

\begin{table*}[!t]
    \caption{Multi-chart parameterization quantitative comparisons of (a) our method, (b) Nuvo \cite{srinivasan2025nuvo}, (c) xatlas\protect\footnotemark[1] and (d) Blender Smart UV\protect\footnotemark[2] in terms of Isometric, Charts number, Seams Length and Optimization Time. The best results are highlighted in \textbf{bold}.} \vspace{-0.3cm}
    \centering
    \renewcommand{\arraystretch}{1.5}
    \begin{tabular}{c|c|c|c|c|c|c|c|c|c|c|c|c|c|c|c|c}
    \toprule[1.5pt]
    ~ & \multicolumn{4}{c|}{Isometric Metric ($\times10^{-4}$)} &  \multicolumn{4}{c|}{\# Charts} &  \multicolumn{4}{c|}{Seams Length}& \multicolumn{4}{c}{Time (Seconds)}\\
    \hline
    Model & (a) & (b)  & (c)  & (d) & (a)&(b)  & (c)  & (d)& (a)&(b)  & (c)  & (d)& (a)&(b)  & (c)  & (d)\\
    \hline 
    Bunny  &6.4&59.6&\textbf{5.6}&20.6
    & \textbf{8}& \textbf{8} & 47 & 497 
    & 29.42 & \textbf{26.97} & 2816 & 2867
    &1371&2261&35&\textbf{0.39}\\
    \hline
    Armadillo &27.3&100&\textbf{10.2}&122
    & \textbf{8}&\textbf{8} & 387 &1235 
    & 35.83 & \textbf{30.22} & 1939 & 2039
    &1313&	2298&	7&	\textbf{0.27}\\
    \hline
    Head &22.1&81.8&\textbf{11.0}&165
    & \textbf{8}&\textbf{8} & 23  &286 
    & \textbf{26.41} & 31.20 & 2089 & 2125
    &1231	&2207	&15	&\textbf{0.16}\\
    \hline
    Monster &\textbf{3.4}&89.2&5.8&27.3
    & \textbf{8}&\textbf{8} & 517  & 599 
    & 41.77 & \textbf{32.07} & 2388 & 2484
    &1287&2189&7&\textbf{0.27}\\
    \hline
    Bird  &30.5&118&\textbf{13.6}&156
    & \textbf{7}& 8 & 40  &85 
    & \textbf{25.11} & 35.81 & 1566 & 1604
    &1309	&2222	&7	&\textbf{0.07}\\
    \hline
    Duck &12.8&81.1&\textbf{6.1}&144
    & \textbf{7}& 8 & 399 &215 
    & \textbf{27.44} & 38.52 & 2431 & 2503
    &1252	&2193	&4	&\textbf{0.24}\\
    \hline
    Hand &\textbf{27.8}&116&33.3&202
    & \textbf{7}& 8 & 25  &75 
    & 36.08 & \textbf{35.39} & 1569 & 1605
    &1224	&2188	&10	&\textbf{0.08}\\
    \hline
    Bear  &41.7&131& \textbf{19.6}&199
     & \textbf{7}& 8 & 27  &46 
     & 35.32 & \textbf{31.71} & 1506 & 1539
     &1303	&2266	&4	&\textbf{0.07}\\
     \hline
    Human &\textbf{24.4}&141&59.4&183
    & \textbf{6}& 8 & 173 &196 
    &\textbf{39.39} & 40.94 & 1862 & 1931
   & 1236	&2275	&6	&\textbf{0.11}\\
    \hline
    Desk &24.2&143&\textbf{13.3}&204
    & \textbf{5}& 8 & 25  &27 
    & 36.92 & \textbf{31.14} & 1522 & 1558
    &1386	&2152	&3	&\textbf{0.07}\\
    \hline
    Dolphin &11.6&78.5&\textbf{5.7}&144
    & \textbf{5}& 8 & 23  &86 
    & \textbf{26.17} & 28.84 & 2158 & 2185
    &1292	&2284	&10	&\textbf{0.16}\\
    
    \bottomrule[1.5pt]
    \end{tabular}
    \label{quantitative_comparison_multi_chart}
\end{table*}

\begin{figure*}[!t]
    \centering
    \includegraphics[width=1.0\linewidth]{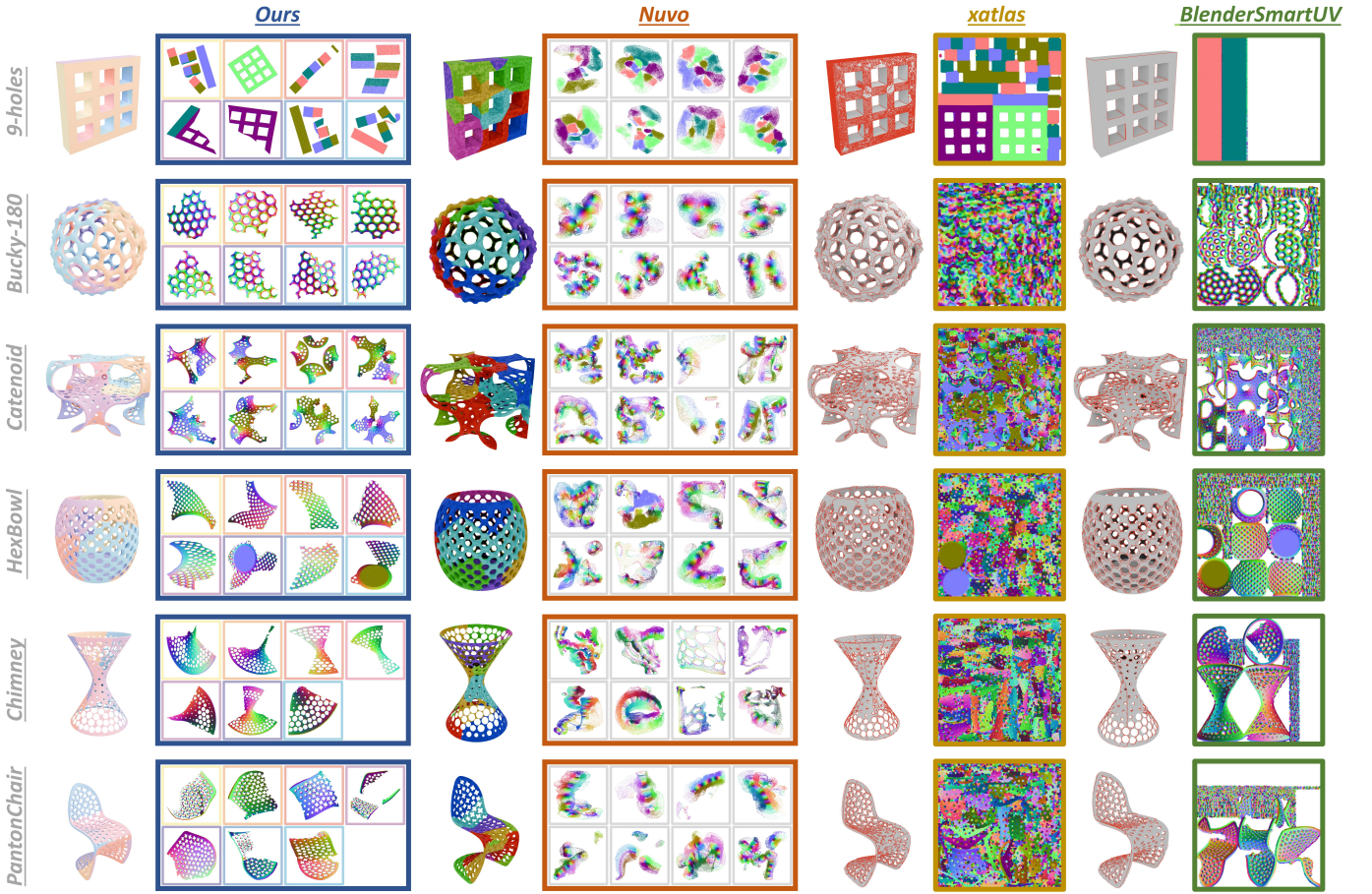}
    \caption{Comparison of multi-chart UV unwrapping results on models with complex topologies. We visualize parameterization results produced by our multi-chart neural surface parameterization method, Nuvo, xatlas and Blender Smart UV. The 2D mapped points are color-coded by surface normals to facilitate visual comparison.}
    \label{visual_comparison_multi_chart_complex}
\end{figure*}
\begin{table*}[!t]
    \caption{Multi-chart parameterization quantitative comparisons of (a) our method, (b) Nuvo \cite{srinivasan2025nuvo}, (c) xatlas\protect\footnotemark[1] and (d) Blender Smart UV\protect\footnotemark[2] on complex models in terms of Isometric, Charts number, Seams Length and Optimization Time. The best results are highlighted in \textbf{bold}.} \vspace{-0.3cm}
    \centering
    \renewcommand{\arraystretch}{1.5}
    \begin{tabular}{c|c|c|c|c|c|c|c|c|c|c|c|c|c|c|c|c}
    \toprule[1.5pt]
    ~ & \multicolumn{4}{c|}{Isometric Metric ($\times10^{-4}$)} &  \multicolumn{4}{c|}{\# Charts} &  \multicolumn{4}{c|}{Seams Length}&\multicolumn{4}{c}{Time (Seconds)}\\
    \hline
       Model & (a) & (b)  & (c)  & (d) & (a)&(b)  & (c)  & (d)& (a)&(b)  & (c)  & (d)& (a)&(b)  & (c)  & (d)\\
    \hline 
    9-holes & 9.4&309&\textbf{4.4}&266
    &\textbf{8}& \textbf{8} & 14561 & 46 
    & \textbf{273.9} & 411.9 & 11670 & 18101
    &1391	&2219	&4	&\textbf{0.41}\\
    \hline
    Bucky-180 &\textbf{10.0}& 1214 & 140&433
    & \textbf{8}& \textbf{8} & 387 &1235 
    & 734.1 & \textbf{554.9} & 8014 & 9889
    &1279	&2189	&3	&\textbf{0.12}\\
    \hline
    Catenoid & \textbf{9.5}& 765 &40.9&237
    & \textbf{8}& \textbf{8} & 23  &286 
    & 728.3 & \textbf{504.3} & 10102 & 12281
    &1373	&2283	&31	&\textbf{0.33}\\
    
    \hline
    HexBowl& 25.6&8021&\textbf{23.5}&98.7
    & \textbf{8}& \textbf{8} & 399 &215 
    & 435.9 & \textbf{268.8} & 11083 & 13223
    &1317	&2206	&12	&\textbf{0.29}\\
    \hline
    Chimney & \textbf{171}&442&335&2340
    & \textbf{7}& 8 & 40  &85 
    & \textbf{473.1} & 483.2 & 12847 & 14373
    &1275	&2186	&73	&\textbf{0.76}\\
    \hline
    PantonChair & \textbf{14.1}&470&32.3&189
    & \textbf{7}& 8 & 22  &272 
    & 646.1 & \textbf{178.1} & 5266 & 6554
    &1255	&2150	&18	&\textbf{0.26}\\
    \bottomrule[1.5pt]
    \end{tabular}
    \label{quantitative_comparison_multi_chart_complex}
\end{table*}

\subsection{Comparison on Multi-Chart Parameterization}

In the multi-chart parameterization study, we compared several widely adopted methods: Nuvo \cite{srinivasan2025nuvo}, xatlas\footnote{J. Young, “xatlas,” https://https://github.com/jpcy/xatlas}, and Blender Smart UV\footnote{B. O. Community, “Blender - a 3D modelling and rendering package,”
http://www.blender.org.}. Unlike global parameterization, multi-chart approaches excel at producing UV unwrapping results with minimal distortion. To this end, our evaluation extends beyond visual comparisons to include isometric distortion, a stricter quantitative metric for surface parameterization.

The number and shape of charts significantly influence the efficacy of multi-chart parameterization. Excessive or overly fragmented charts are undesirable, as they can impair downstream tasks such as texture painting. Therefore, in addition to distortion metrics, we evaluated the number of charts and the total seam length produced by each method.

For xatlas and Blender Smart UV, we used their default configurations. In contrast, both Nuvo and our proposed method allow the specification of a target chart number, which we set uniformly to $n=8$. This parameter choice is based on the observation that a higher $n$ simplifies the parameterization process. For most models, 8 charts are sufficient to achieve low-distortion results. Furthermore, capping the chart number at 8 enables us to evaluate whether a method can produce segmentations that effectively support subsequent parameterization under constrained conditions.

Figure \ref{visual_comparison_multi_chart} presents a visual comparison of parameterization results across a range of common models. When constrained to a limited number of charts, our method produces segmentations with clear geometric significance. In contrast, Nuvo tends to generate clustered, near-circular segmentations due to its cluster loss constraint. While this regularity benefits complex shapes, it significantly limits Nuvo’s parameterization potential, as the resulting segments deviate from approximating developable surfaces. This limitation is particularly evident in \textit{Dolphin}, where Nuvo’s cluster loss fails to favor the elongated, nearly developable side. Meanwhile, xatlas and Blender Smart UV, which do not impose chart number restrictions, often produce overly fragmented segmentations for certain models (e.g., \textit{Monster} and \textit{Duck}). These methods perform adequately for models with flatter geometries, such as \textit{Desk}, but struggle with highly curved surfaces, as observed in \textit{Armadillo}. Notably, although constrained to $n=8$, our method adaptively reduces and merges charts for simpler models when distortion is sufficiently controlled, as seen in \textit{Bird}, \textit{Desk}, and \textit{Dolphin}.

Quantitative results, as presented in Table \ref{quantitative_comparison_multi_chart}, reinforce these findings. In terms of distortion, Nuvo and Blender Smart UV perform less effectively than our method and xatlas. Both our approach and xatlas consistently achieve optimal or near-optimal distortion metrics across all models. However, xatlas’s overly fragmented segmentations and longer seam lengths make its results impractical for certain models (e.g., \textit{Duck}: 399 charts, \textit{Armadillo}: 387 charts). Nuvo achieves a better balance between the chart number and seam length, albeit at the cost of higher distortion. Overall, our method offers the most favorable trade-off among charts number, seam length, and distortion.

\begin{figure}[!t]
    \centering
    \includegraphics[width=1.0\linewidth]{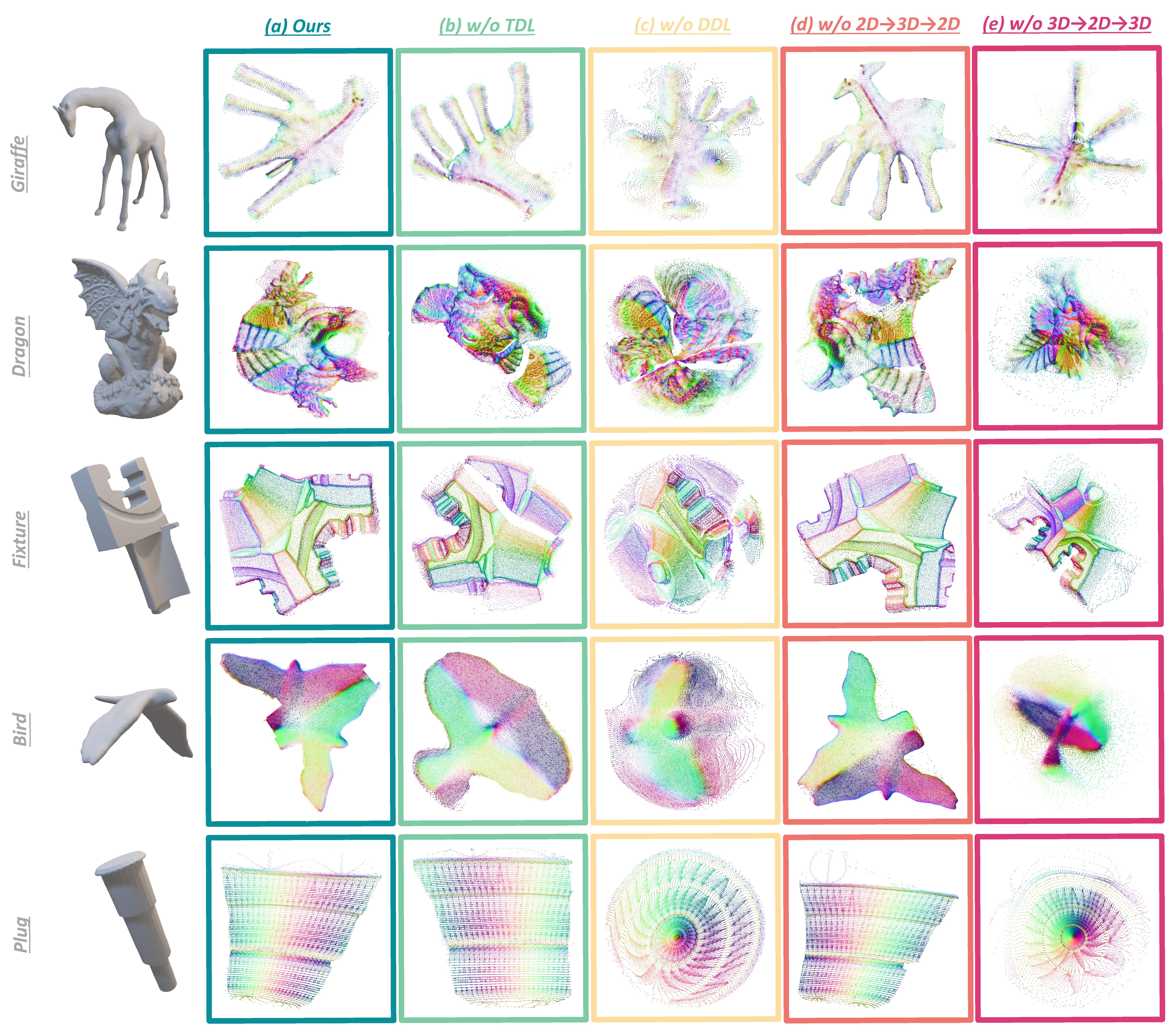}
    \caption{Ablation study showing visual results generated by: (a) our full framework, (b) the model without TDL, (c) without DDL, (d) without the upper 2D$\rightarrow$ 3D$\rightarrow$ 2D branch, and (e) without the bottom 3D$\rightarrow$ 2D$\rightarrow$ 3D branch.}
    \label{global_visual_ablation}
\end{figure}
\begin{table}[!t]
    \caption{Ablation quantitative results on different configurations of global parameterization produced by (a) our full framework, (b) without TDL, (c) without DDL, (d) without upper 2D$\rightarrow$3D$\rightarrow$2D branch and (e) without bottom 3D$\rightarrow$2D$\rightarrow$3D branch. The best results are highlighted in \textbf{bold}.} \vspace{-0.3cm}
    \centering
    \renewcommand{\arraystretch}{1.5}
    \begin{tabular}{c|c|c|c|c|c}
    \toprule[1.5pt]
    Model & (a) & (b) & (c)  &  (d) & (e) \\
    \hline
    Giraffe &  \textbf{0.071} & 0.088 & 0.424 & 0.075 & 0.209\\
    \hline
    Dragon  &  \textbf{0.123} & 0.137 & 0.287  & 0.139 & 0.193\\
    \hline
    Fixture  &  \textbf{0.057} & 0.090 & 0.387  & 0.083 & 0.158\\
    \hline
    Bird &  \textbf{0.025} & 0.034 & 0.280  & 0.027 & 0.129\\
    \hline
    Plug &  \textbf{0.015}& 0.016 & 0.228 & 0.022 & 0.150\\
    \bottomrule[1.5pt]
    \end{tabular}
    \label{global_quantitative_ablation}
\end{table}

\begin{figure}[!t]
    \centering
    \includegraphics[width=0.9\linewidth]{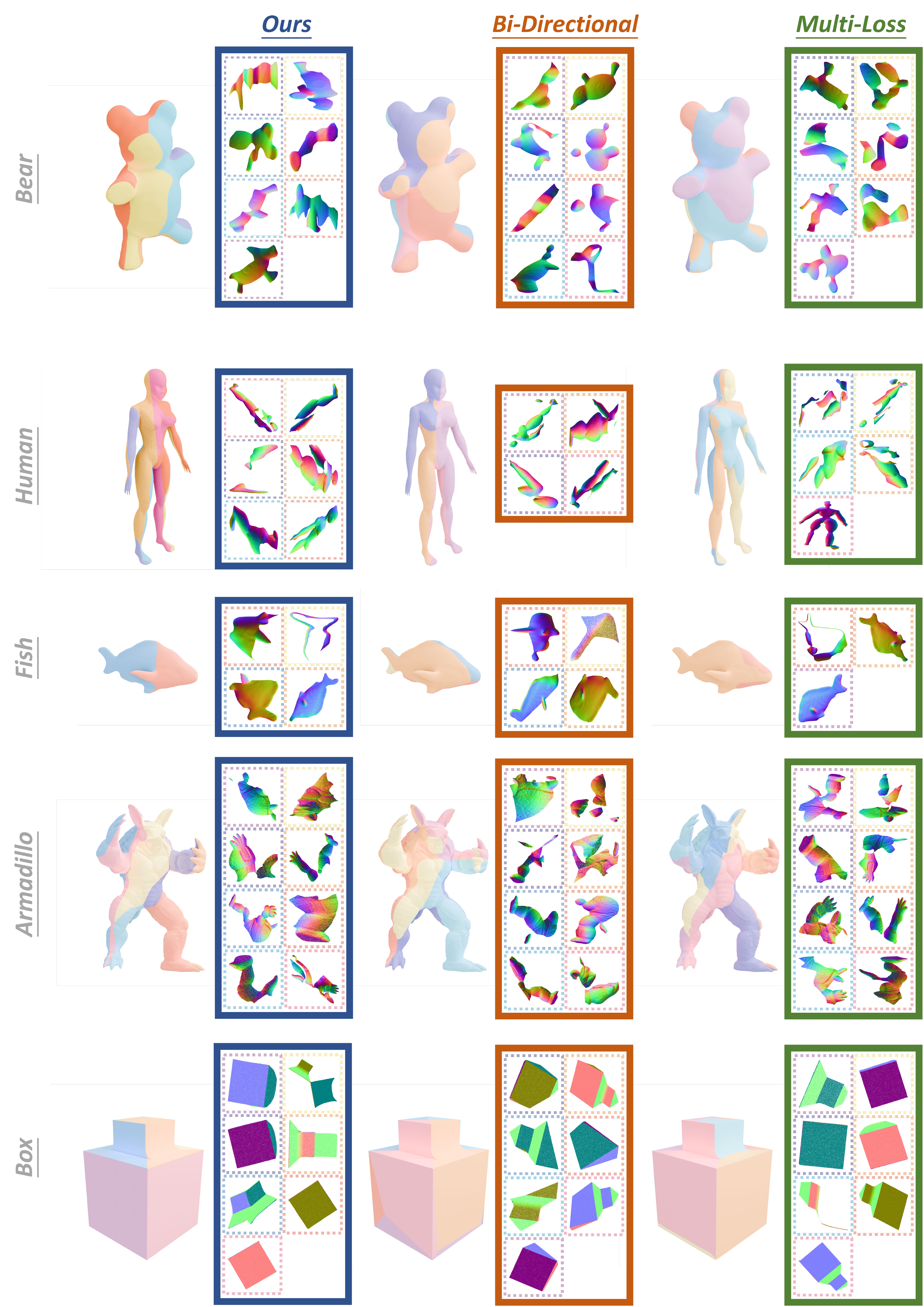}
    \caption{Ablation results for multi-chart parameterization, generated by our full framework, the model incorporating bi-directional branches and with Multi-Loss by Triangle Distortion and Differential Distortion.}
    \label{multi_chart_visual_ablation}
\end{figure}
\begin{table*}[!t]
    \caption{Ablation quantitative results in multi-chart parameterization produced by our framework, with Bi-Directional branches and with Multi-Loss by Triangle Distortion and Differential Distortion in terms of Conformal, Equiareal, Isometric and Optimization Time costs (in Hours).} \vspace{-0.3cm}
    \centering
    \renewcommand{\arraystretch}{1.5}
    \begin{tabular}{c|c|c|c|c|c|c|c|c|c|c|c|c}
    \toprule[1.5pt]
    ~ & \multicolumn{3}{c|}{Conformal Metric~($\times10^{-3}$)} &  \multicolumn{3}{c|}{Equiareal  Metric~($\times10^{-6}$)} &  \multicolumn{3}{c}{Isometric Metric~($\times10^{-4}$)}&\multicolumn{3}{|c}{Time~(Hours)}\\
    \hline
    ~& Ours & Bi-Direct. & Multi-Loss & Ours & Bi-Drect. & Multi-Loss& Ours & Bi-Direct. & Multi-Loss & Ours & Bi-Direct. & Multi-Loss\\
    \hline 
    Bear & 14.4& 10.7 & 12.2 
    &9.22 & 5.86 & 5.64
    &41.7& 61.4 & 24.3
    & 0.34 & 1.53 & 8.95\\
    \hline
    Human & 51.8 & 37.3 & 49.3
    &4.41 & 3.23 & 4.26
    &24.4 & 37.3 & 23.4
    & 0.36 & 1.51 & 9.19\\
    \hline
    Fish & 33.6 & 45.9 & 37.7  
    & 3.59 & 2.18 & 2.48
    & 21.4 & 52.3 & 14.6
    & 0.45 & 1.57 & 9.21\\
    \hline
    Armadillo & 52.4 & 29.1 & 40.7 
    &3.52 & 1.61 & 4.87
    &20.8 & 34.2 & 30.4
    & 0.37 & 1.50 & 9.36\\
    \hline
    Box & 3.7 & 7.3 & 7.3 
    & 5.00 & 6.46 & 4.33
    & 23.8 & 50.8 & 21.2 
    & 0.35  & 1.56 & 8.99\\
    \bottomrule[1.5pt]
    \end{tabular}
    \label{multi_chart_quantitative_ablation}
\end{table*}
\begin{figure*}[!t]
    \centering
    \includegraphics[width=0.85\linewidth]{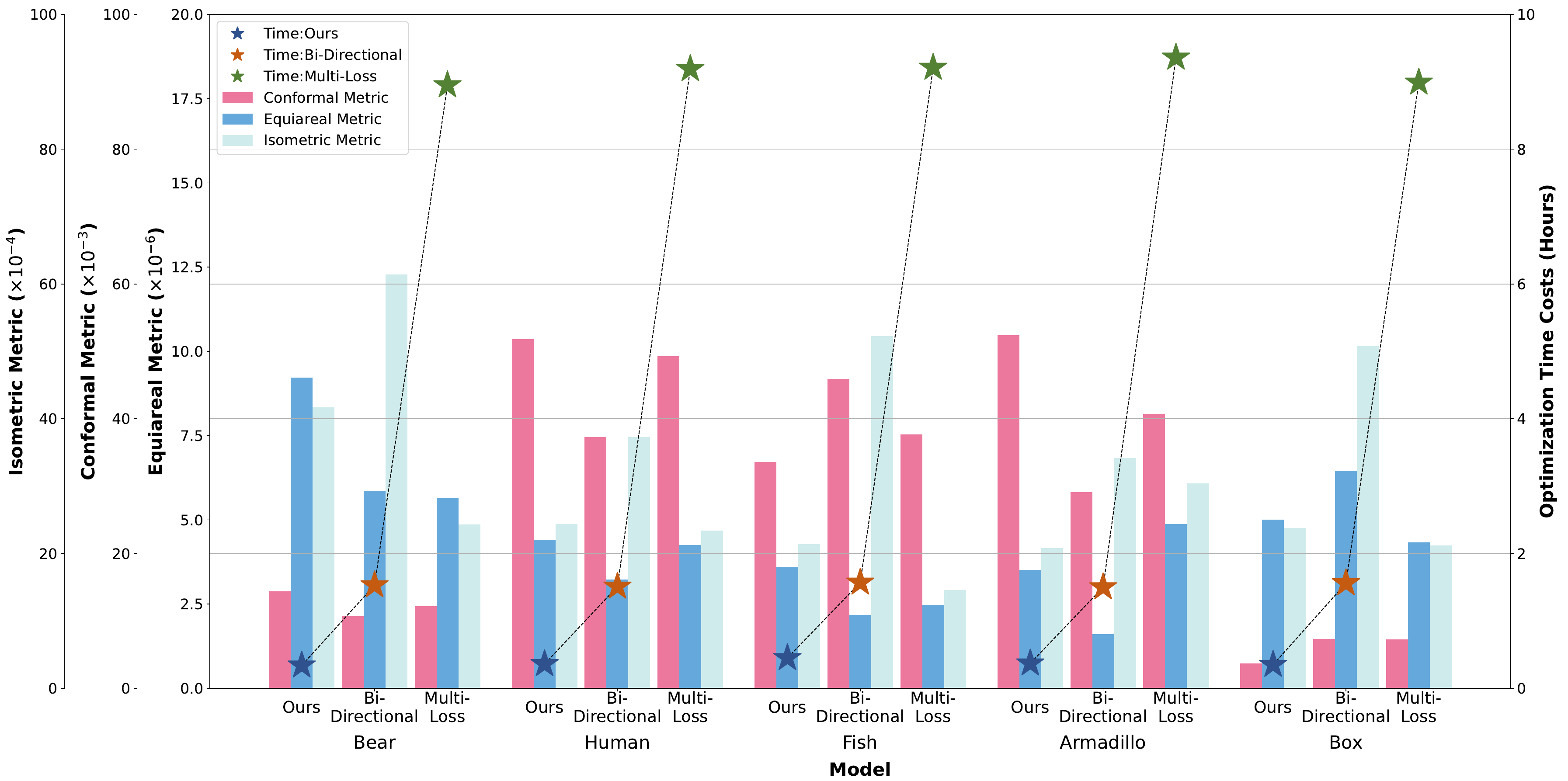}
    \caption{Comparison of optimization time costs (Hours) and distortion of different configurations of multi-chart parameterization.}
    \label{comparison_time_cost}
\end{figure*}
To further demonstrate the robustness of our method, we conducted stress tests on models with high genus and intricate topologies. Figure \ref{visual_comparison_multi_chart_complex} presents the visual results, and Table \ref{quantitative_comparison_multi_chart_complex} summarizes the quantitative results. For the highly regular \textit{9-holes} model, xatlas delivers exceptional segmentations, whereas Blender Smart UV performs poorly. Our method produces commendable segmentations that closely align with the model’s sharp edges. Nuvo, constrained by its clustering loss, is less sensitive to such edges, resulting in suboptimal segmentations. Across the remaining five models, our method consistently achieves satisfactory parameterization results. Nuvo’s segmentations are acceptable but exhibit deficient distortion control. Xatlas struggles to produce meaningful results for these complex models, while Blender Smart UV, though capable of some coherent segmentations, generates excessively fragmented outcomes globally.\footnotetext[1]{J. Young, “xatlas,” https://https://github.com/jpcy/xatlas}\footnotetext[2]{B. O. Community, “Blender - a 3D modelling and rendering package,”
http://www.blender.org.} Table \ref{quantitative_comparison_multi_chart} and Table \ref{quantitative_comparison_multi_chart_complex} also compare optimization times. Xatlas and Blender Smart UV demonstrate high efficiency, but their UV results are overly fragmented. Nuvo, reliant on neural radiance fields, requires a longer training process in multi-chart tasks than our method. In conclusion, our analysis provides the following insights:
\footnotetext[1]{J. Young, “xatlas,” https://https://github.com/jpcy/xatlas}\footnotetext[2]{B. O. Community, “Blender - a 3D modelling and rendering package,”
http://www.blender.org.}
\begin{itemize}
    
    \item Under the constraint of a limited charts number, our method achieves significantly low parameterization distortion. Compared to Nuvo, which also limits chart numbers, our approach exhibits markedly superior low-distortion performance. Relative to xatlas, which similarly achieves minimal distortion, our method’s reduced charts number enhances its practical applicability.
    
    \item For models with high genus and complex topologies, our method shows no notable performance degradation, consistently producing meaningful multi-chart parameterization results.
    
\end{itemize}

\subsection{Ablation Study}
\subsubsection{Global Parameterization}

In the context of global parameterization, we conducted ablation studies to evaluate the contributions of our bi-directional cycle mapping architecture and distortion loss functions. Specifically, we tested the following configurations: (a) our complete framework, (b) excluding the TDL, (c) excluding the DDL, (d) excluding the upper 2D$\rightarrow$3D$\rightarrow$2D branch, and (e) excluding the lower 3D$\rightarrow$2D$\rightarrow$3D branch. Visual results of these ablation experiments are presented in Figure \ref{global_visual_ablation}, and quantitative conformal distortion metrics for each configuration are reported in Table \ref{global_quantitative_ablation}.

The DDL is the primary component of our distortion loss functions, playing a more critical role in constraining distortion. Without DDL, relying solely on the TDL—which constrains only discrete mesh vertices—leads to insufficient global stability. This limitation disrupts the balance between local and global distortion control, consequently impairing the ability to identify optimal seams. Conversely, when using DDL alone, the parameterization achieves favorable global results but produces some outliers. The inclusion of TDL mitigates these outliers to some extent, reducing conformal distortion, as evidenced by the quantitative results in Table \ref{global_quantitative_ablation}.

The bi-directional cycle mapping architecture achieves the lowest conformal distortion. The configuration with only the 3D$\rightarrow$2D$\rightarrow$3D branch, excluding the 2D$\rightarrow$3D$\rightarrow$2D branch, yields reasonably good results but exhibits instability. Incorporating the 2D$\rightarrow$3D$\rightarrow$2D branch significantly enhances the stability of seam identification and parameterization. In contrast, the configuration with only the 2D$\rightarrow$3D$\rightarrow$2D branch, which serves as an auxiliary branch, struggles to achieve satisfactory parameterization performance when used independently.

\subsubsection{Multi-Chart Parameterization}

In the multi-chart parameterization framework, we employed only the 3D$\rightarrow$2D$\rightarrow$3D branch and the TDL from the global parameterization architecture. This design choice is driven by the fact that $n$-chart parameterization requires executing $n$ neural parameterization processes simultaneously, increasing computational time by a factor of $n$ compared to global parameterization. Additionally, the bi-directional cycle mapping structure and complex distortion loss functions in global parameterization are primarily designed to enhance seam identification, a task largely delegated to the chart-assignment module in multi-chart parameterization. Theoretically, each independent chart is well-suited for planar parameterization. Thus, this ablation study aims to validate the preservation of low distortion and the reduction in training time under a simplified neural surface parameterization scheme.

We evaluated three multi-chart parameterization configurations using five models: (a) our full framework, (b) with an added bi-directional branch, and (c) with DDL constraints. Figure \ref{multi_chart_visual_ablation} demonstrates robust performance across all configurations. The distortion metrics reported in Table \ref{multi_chart_quantitative_ablation} further corroborate these findings. Additionally, Table \ref{multi_chart_quantitative_ablation} compares training times, revealing that the bi-directional configuration requires approximately 3.5 times the training duration of our full framework, while the DDL configuration demands roughly 20 times longer. These results highlight that our approach achieves significant efficiency improvements with minimal compromise to low-distortion performance. Figure \ref{comparison_time_cost} visually illustrates the comparable low-distortion outcomes and varying efficiencies across the five models under different configurations.
\revise{
\subsection{Others}
\subsubsection{Scalability}
As a neural network-based method, the efficiency of our approach significantly benefits from improvements in GPU performance. Efficiency is particularly crucial when processing large datasets. We tested the operational efficiency of our method on three GPUs: RTX3090, RTX5090, and H800. The results, as shown in Figure \ref{device_time}, indicate that compared to the RTX3090, the RTX5090 improves efficiency by approximately 20\%, while the H800 improves efficiency by about 75\%. The enhanced operational efficiency brought by GPU performance enables our method to scale effectively on large datasets.
\begin{figure}[!t]
    \centering
    \includegraphics[width=0.8\linewidth]{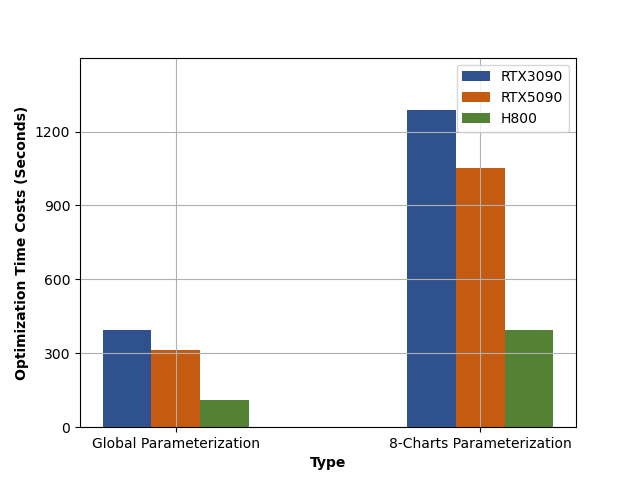}
    \caption{\revise{Comparison of runtime for different GPUs.}}
    \label{device_time}
\end{figure}
\subsubsection{Real Texture Mesh from Real World Scans}
In addition to synthetic data resembling artificial models, our method is equally applicable to models with real textures obtained from real-world scans. We selected two examples to illustrate this point. The first is a scene dataset, sourced from the Replica dataset \cite{replica19arxiv}, also scanned from real-world scenes (Figure \ref{scene}). The second is the bird-house-model, sourced from the DTC dataset \cite{Dong_2025_CVPR}, scanned from real-world objects (Figure \ref{global-bird-house} and Figure \ref{multi-chart-bird-house}).
\begin{figure}[!t]
    \centering
    \includegraphics[width=0.7\linewidth]{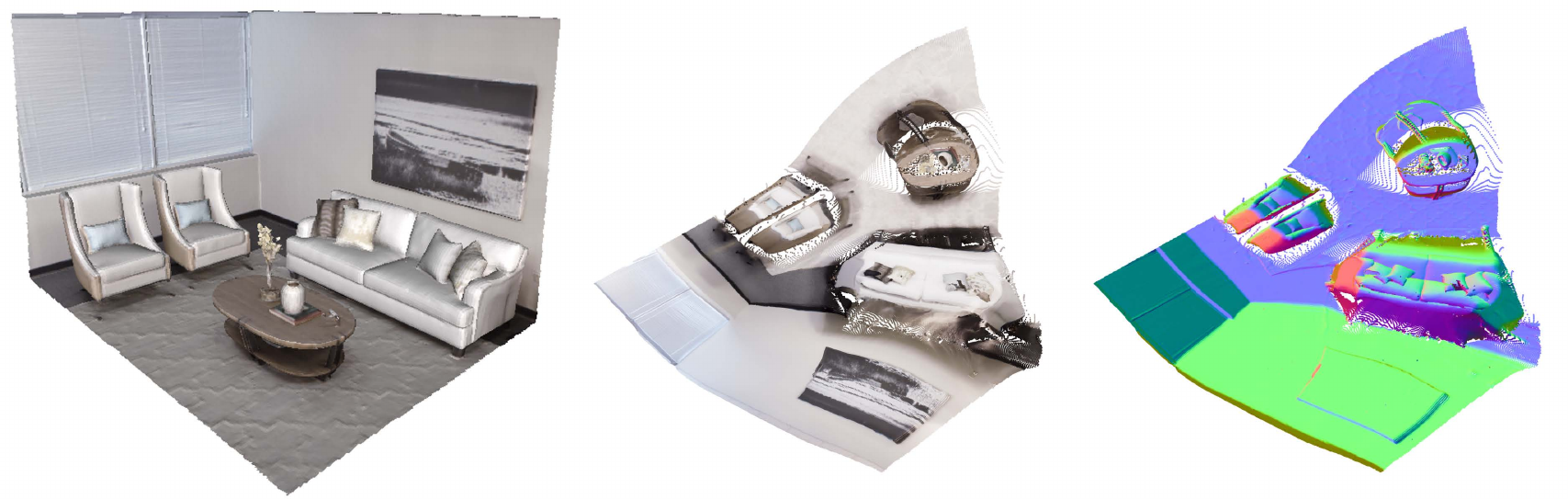}
    \caption{\revise{Parameterization results of real scenes.}}
    \label{scene}
\end{figure}
\begin{figure}[!t]
    \centering
    \includegraphics[width=0.7\linewidth]{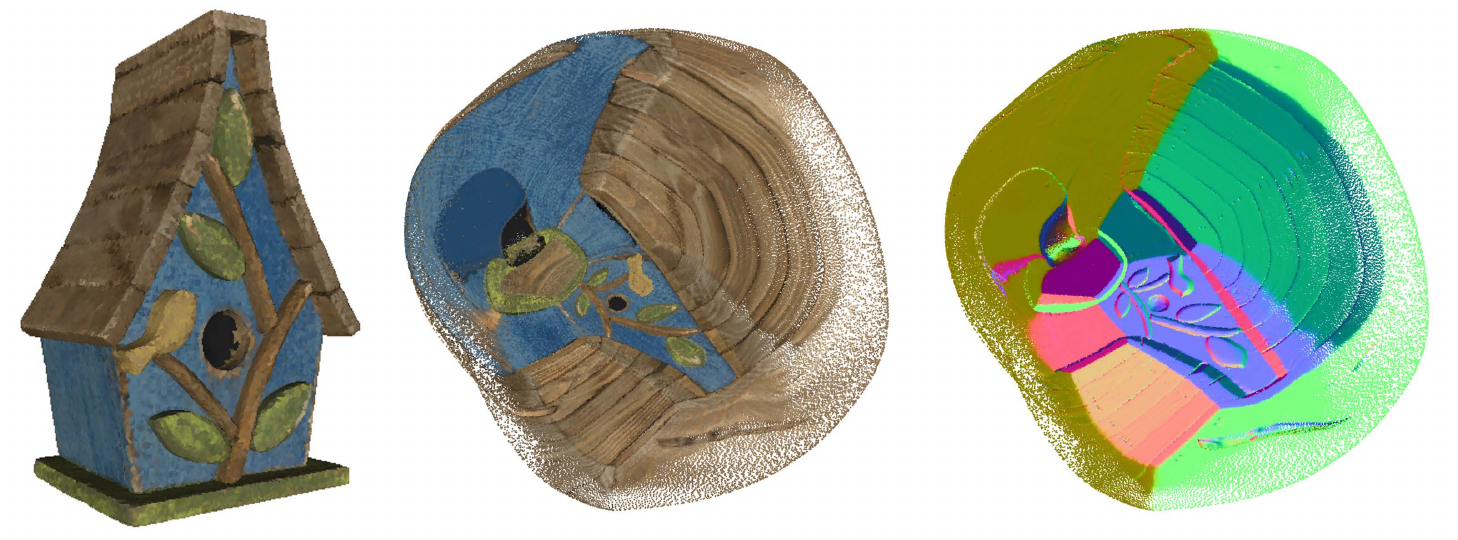}
    \caption{\revise{Global parameterization results of real objects.}}
    \label{global-bird-house}
\end{figure}
\begin{figure}[!t]
    \centering
    \includegraphics[width=0.7\linewidth]{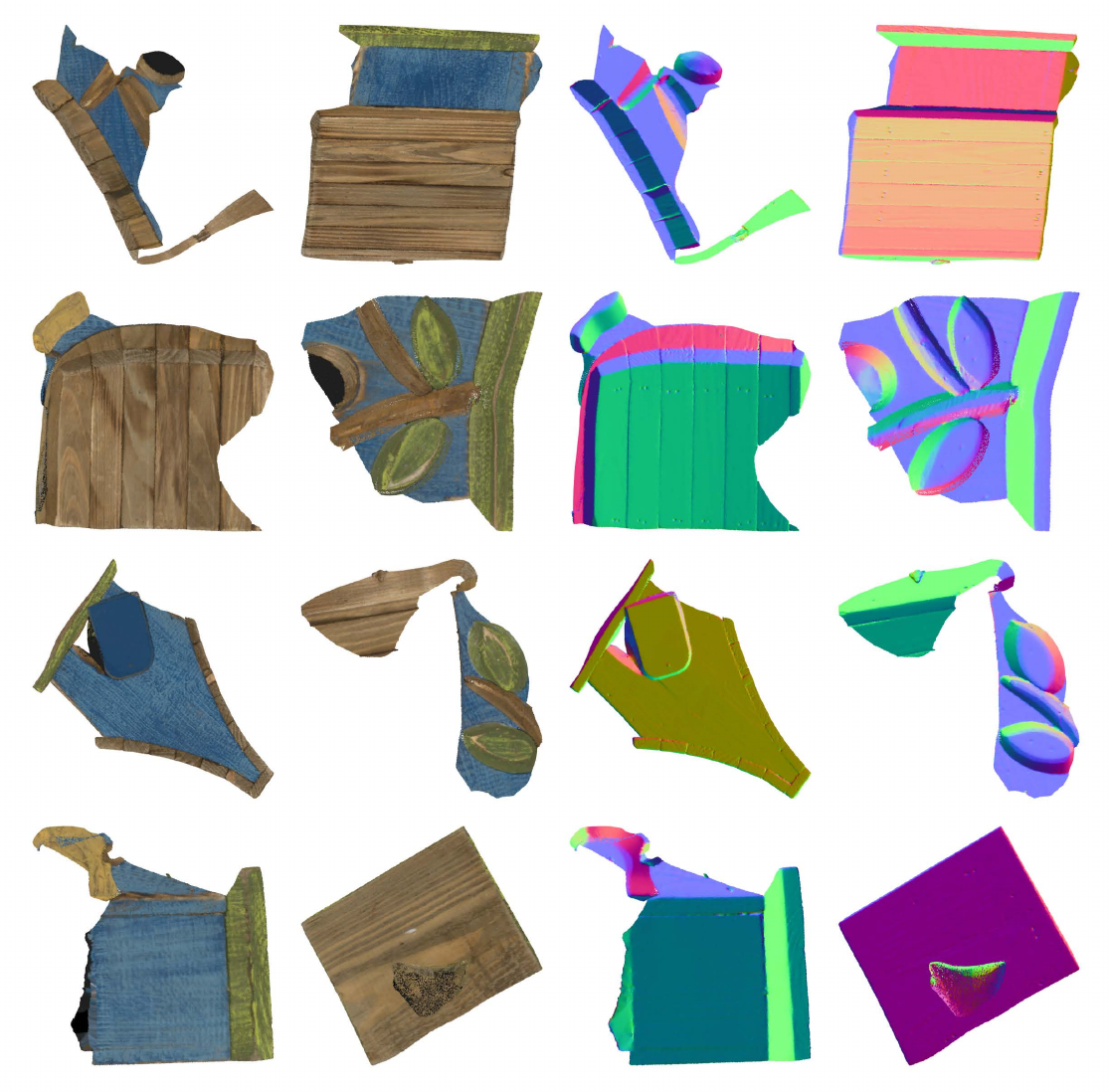}
    \caption{\revise{8-charts parameterization results of real objects.}}
    \label{multi-chart-bird-house}
\end{figure}
\subsubsection{Texture Mapping Application}
We chose texture mapping to demonstrate the potential of our method in downstream task applications. In Figure \ref{application}, we used the bunny model to demonstrate global texture mapping supported by global parameterization and the box model to showcase per-chart texture mapping.
\begin{figure}[!t]
	\centering
	\includegraphics[width=0.7\linewidth]{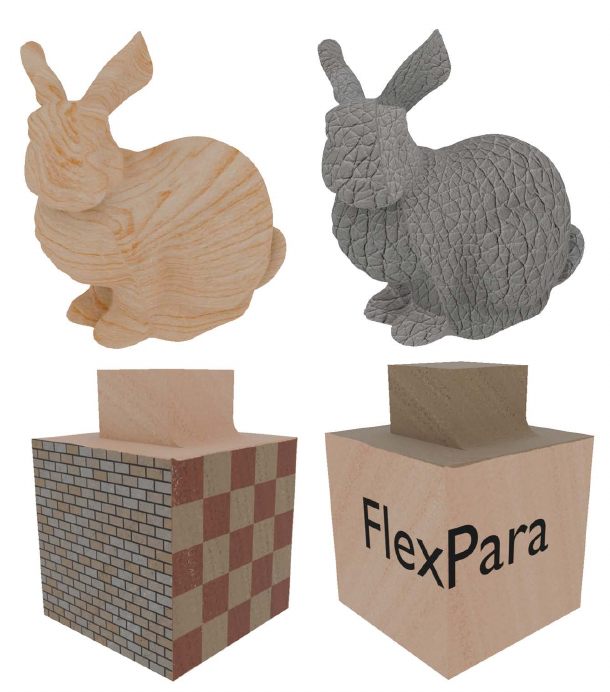}
	\caption{\revise{Texture mapping application results for global parameterization (bunny) and multi-chart parameterization (box).}}
	\label{application}
\end{figure}
}
\section{Conclusion} \label{Conclusion}
This paper presents FlexPara, an unsupervised neural optimization framework for global and multi-chart surface parameterization. For global parameterization, we introduced the first neural surface parameterization method designed for global free-boundary parameterization, which autonomously identifies optimal cutting seams on the target 3D surface while adaptively adjusting the 2D UV parameter domain. Leveraging the efficiency and accuracy of global parameterization in determining cutting seams and UV mappings, we developed a multi-chart neural parameterization framework. This framework employs multiple single-directional cycle mappings to collaboratively optimize chart assignments, achieving superior parameterization quality.

Experimental results on a diverse set of 3D models demonstrate that our method delivers state-of-the-art performance, particularly on challenging models with complex topologies and thin structures.

Given the effectiveness and versatility of our framework, we believe it has significant potential for a wide range of downstream applications. In future work, we plan to explore its applications in tasks such as compressed representation \cite{zeng2024dynamic}, regularized representation \cite{zhang2023flattening}, and the editing and generation of 3D assets. \revise{To enhance the scalability of our method, we plan to explore feedforward methods based on large-scale models in the future to replace the current overfitting framework, further improving scalability.}

\bibliographystyle{IEEEtran}
\bibliography{bib}
\begin{IEEEbiography}[{\includegraphics[width=1in,height=1.25in,clip,keepaspectratio]{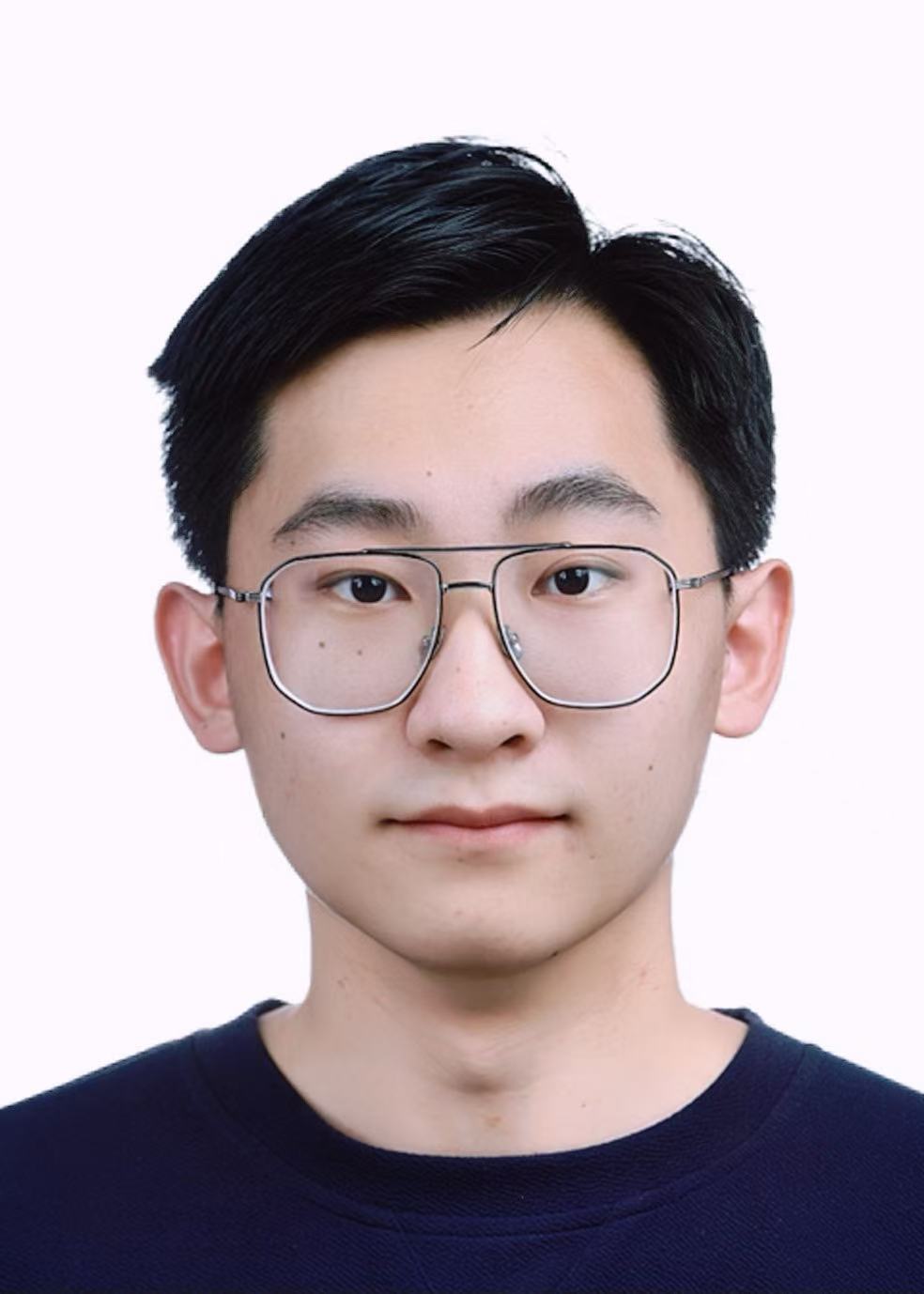}}]{Yuming Zhao} received the B.S. degree in Computer Science and Technology from Beijing Normal University in 2021, and the M.Eng. degree in Computer Applied Technology from Beijing Normal University in 2024. He is currently working toward a PhD degree with the Department of Computer Science, City University of Hong Kong. His research interests include computer graphics, geometry processing, and 3D generative models.
\end{IEEEbiography}

\begin{IEEEbiography}[{\includegraphics[width=1in,height=1.25in,clip,keepaspectratio]{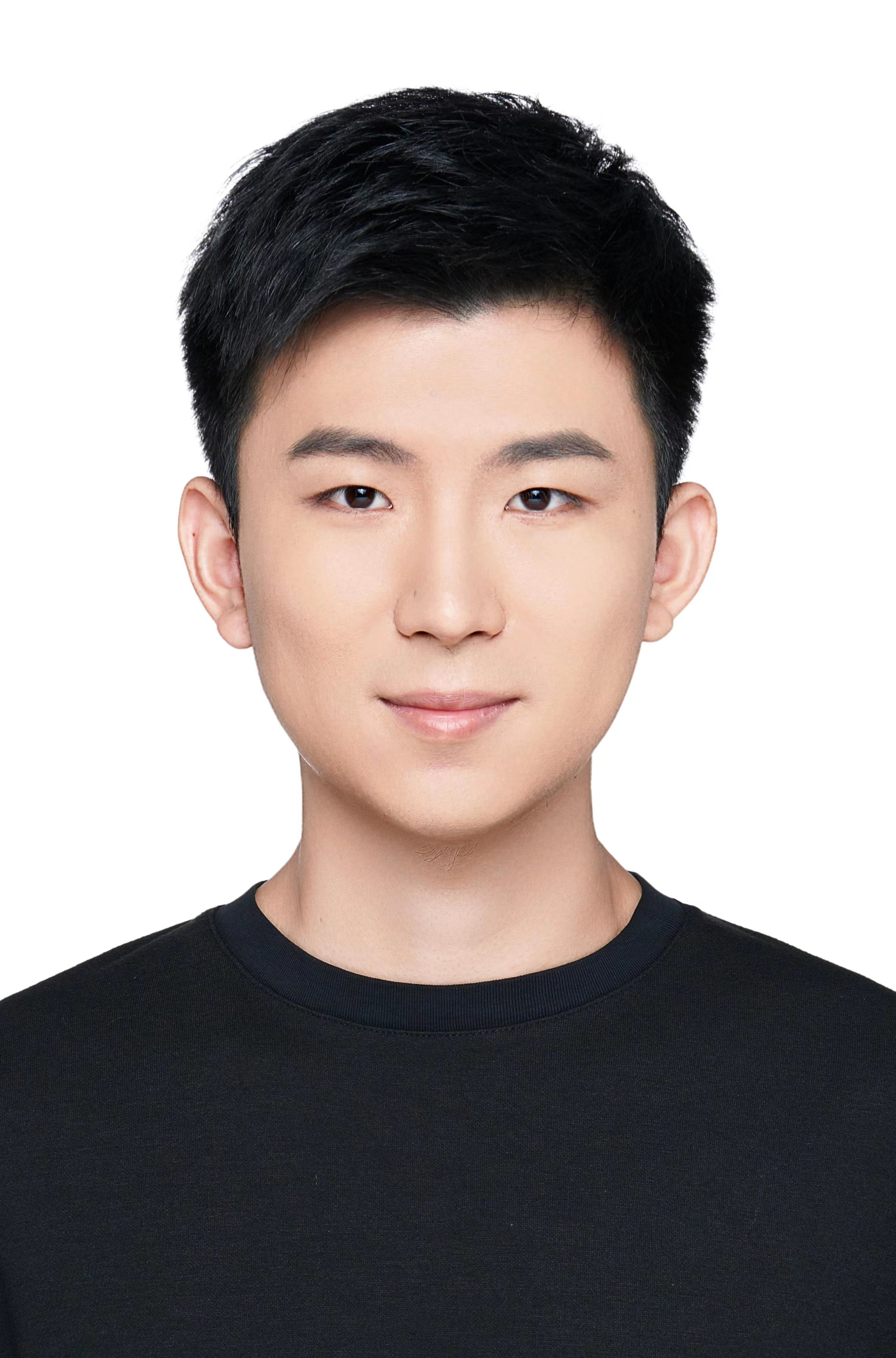}}]{Qijian Zhang} received the B.S. degree in Electronic Information Science and Technology from Beijing Normal University, Beijing, China, in 2019, and then the Ph.D. degree in Computer Science from City University of Hong Kong, Hong Kong SAR, in 2024. He is currently an applied researcher in the TiMi L1 Studio of Tencent Games, focusing on 3D generative models and AI for games.
\end{IEEEbiography}

\begin{IEEEbiography}[{\includegraphics[width=1in, height=1.25in, clip, keepaspectratio]{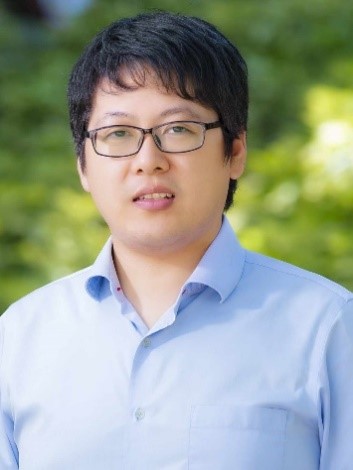}}]{Junhui Hou} (Senior Member, IEEE) is an Associate Professor with the Department of Computer Science, City University of Hong Kong. He holds a B.Eng. degree in information engineering (Talented Students Program) from the South China University of Technology, Guangzhou, China, an M.Eng. degree in signal and information processing from Northwestern Polytechnical University, Xi’an, China, and a Ph.D. degree from the School of Electrical and Electronic Engineering, Nanyang Technological University, Singapore. His research interests are multi-dimensional visual computing.

Dr. Hou received the Early Career Award from the Hong Kong Research Grants Council in 2018 and the NSFC Excellent Young Scientists Fund in 2024. He has served or is serving as an Associate Editor for \textit{IEEE Transactions on Visualization and Computer Graphics}, \textit{IEEE Transactions on Image Processing}, \textit{IEEE Transactions on Multimedia}, and \textit{IEEE Transactions on Circuits and Systems for Video Technology}.  
\end{IEEEbiography}

\begin{IEEEbiography}[{\includegraphics[width=1in, height=1.25in, clip, keepaspectratio]{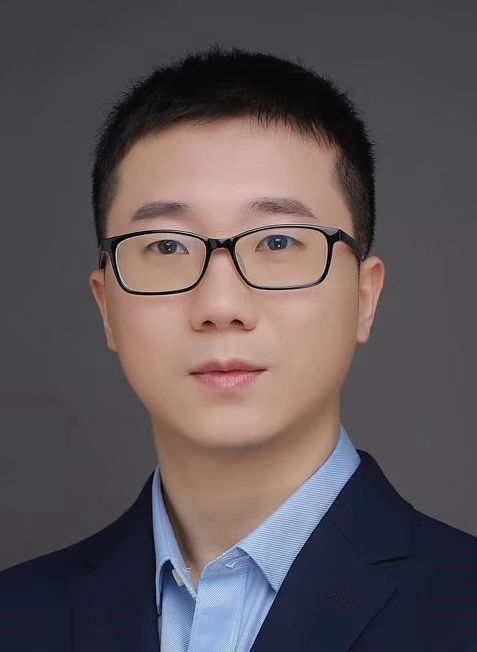}}]{Jiazhi Xia} is a Professor in the School of Computer Science and Engineering, Central South University. His research interests include visualization, geometry processing, and 3D vision. 
\end{IEEEbiography}

\begin{IEEEbiography}[{\includegraphics[width=1in, height=1.25in, clip, keepaspectratio]{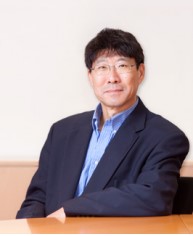}}]
{Wenping Wang} (Fellow, IEEE) is a Professor of Computer Science and Engineering at Texas A\&M University. He works in the areas of computer graphics, computer vision, and geometric computing, and has published extensively in these areas. He received the John Gregory Memorial Award, the Tosiyasu Kunii Award, and the Bézier Award for contributions in geometric computing and shape modeling. He is an ACM Fellow and an IEEE Fellow.
\end{IEEEbiography}

\begin{IEEEbiography}[{\includegraphics[width=1in, height=1.25in, clip, keepaspectratio]{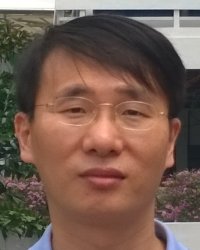}}]{Ying He} is an Associate Professor at the School of Computer Engineering, Nanyang Technological University, Singapore. His research interests are primarily in geometric computing and analysis. He actively participates in the technical program committees of major conferences in geometric modeling and is serving/has served on the editorial boards of IEEE Transactions on Visualization and Computer Graphics, Computer Graphics Forum, and Computational Visual Media. He has also served as General/Program Co-Chair for the Shape Modeling International conference in 2022, the Symposium on Solid and Physical Modeling in 2022 and 2023, the Geometric Modeling and Processing conference in 2014 and 2021, and the Conference on Computational Visual Media in 2020.
\end{IEEEbiography}

\end{document}